\documentclass{article}

\usepackage[final]{neurips_2025}

\usepackage[utf8]{inputenc} 
\usepackage[T1]{fontenc}    
\usepackage{hyperref}       
\usepackage{url}            
\usepackage{booktabs}       
\usepackage{amsfonts}       
\usepackage{nicefrac}       
\usepackage{microtype}      
\usepackage{xcolor}         

\usepackage{algorithm}
\usepackage{algcompatible}
\usepackage{multirow}
\usepackage{graphicx}
\usepackage{amsmath}
\usepackage{amssymb}
\usepackage{amsfonts}       
\usepackage{booktabs}
\usepackage{adjustbox}
\usepackage{hyperref}
\usepackage{xcolor}
\usepackage{nccmath}
\usepackage{wrapfig}  
\usepackage{xspace}
\usepackage{spverbatim}
\usepackage{subcaption}
\usepackage{wrapfig}
\usepackage{minitoc}
\usepackage{tikz}

\newcommand{\sysn}{\text{SamS}\xspace}

\title{
Adaptive Batch-Wise Sample Scheduling for Direct Preference Optimization
}

\author{%
  Zixuan Huang${}^{1,2}$ \ \ \ \ 
  Yikun Ban${}^{1}\thanks{Deqing Wang and Yikun Ban are corresponding authors.} $  \ \ \ \ 
  Lean Fu${}^{2}$\ \ \ \ 
  Xiaojie Li${}^{1}$ \\ 
  \textbf{Zhongxiang Dai}${}^{3}$ \ \ \ \ 
  \textbf{Jianxin Li}${}^{1}$ \ \ \ \ 
  \textbf{Deqing Wang${}^{1*}$} \\
  ${}^{1}$Beihang University \ \ \ 
  ${}^{2}$Bytedance China \ \ \ 
  ${}^{3}$The Chinese University of Hong Kong, Shenzhen \\
  \texttt{\{huang\_zx, yikunb, li\_xiaojie, dqwang\}@buaa.edu.cn} \\
  \texttt{lijx@act.buaa.edu.cn} \ \ \
  \texttt{fulean@bytedance.com} \ \ \ 
  \texttt{daizhongxiang@cuhk.edu.cn}
}

\begin{document}

\maketitle


\begin{abstract}
Direct Preference Optimization (DPO) has emerged as an effective approach for aligning large language models (LLMs) with human preferences. However, its performance is highly dependent on the quality of the underlying human preference data. To address this bottleneck, prior work has explored various data selection strategies,  but these methods often overlook the impact of the evolving states of the language model during the optimization process.
In this paper, we introduce a novel problem: Sample Scheduling for DPO, which aims to dynamically and adaptively schedule training samples based on the model's evolving batch-wise states throughout preference optimization. To solve this problem, we propose SamS, an efficient and effective algorithm that adaptively selects samples in each training batch based on the LLM's learning feedback to maximize the potential generalization performance.
Notably, without modifying the core DPO algorithm, simply integrating SamS significantly improves performance across tasks, with minimal additional computational overhead. 
This work points to a promising new direction for improving LLM alignment through batch-wise sample selection, with potential generalization to RLHF and broader supervised learning paradigms.
The code is available at \url{https://github.com/hzx122/SamS}.

\end{abstract}

\addtocontents{toc}{\protect\setcounter{tocdepth}{-1}}

\section{Introdcution}  \label{intro}

Direct Preference Optimization (DPO)~\cite{rafailov2023direct} was proposed as a simpler and more stable alternative to Reinforcement Learning from Human Feedback (RLHF)~\cite{christiano2017deep,ziegler2019fine,ouyang2022training,  glaese2022improving,chakraborty2024maxmin}. As an off-policy preference optimization method, DPO does not require first training an explicit reward model. Instead, given a preference dataset where each sample includes a prompt and a pair of generations with the first one more consistent with human preferences, it directly optimizes a straightforward binary cross-entropy-type objective, which increases the likelihood of chosen response and decreases the likelihood of rejected response. The promise of this approach is that it implicitly optimizes the same objective as RLHF without adding complexity.

Although DPO has demonstrated exceptional performance across a wide range of tasks, its heavy reliance on high-quality human preference data poses a significant bottleneck for practical deployment due to the associated annotation costs. To mitigate this challenge, substantial research efforts have been devoted to enhancing the data quality and utilization in preference optimization. These efforts generally fall into three categories:
(1) Active Querying~\cite{das2024active, muldrew2024active, ji2024reinforcement}: selecting informative samples for human feedback collection;
(2) Response Pair Selection~\cite{mehta2023sample, liu2024sample}: actively choosing response pairs to annotate conditioned on a given query;
(3) Data Pre-selection~\cite{shen2024towards, deng2025less, gao2025principled}: identifying and filtering high-quality samples prior to DPO training.
However, approaches in categories (1) and (2) typically only focus on online feedback collection and ignore data quality, while methods in category (3) overlook the evolving internal states of the language model throughout the DPO process. 

\begin{figure}[ht]
    \centering
    \begin{subfigure}[b]{0.49\textwidth}
        \centering
        \includegraphics[width=\textwidth]{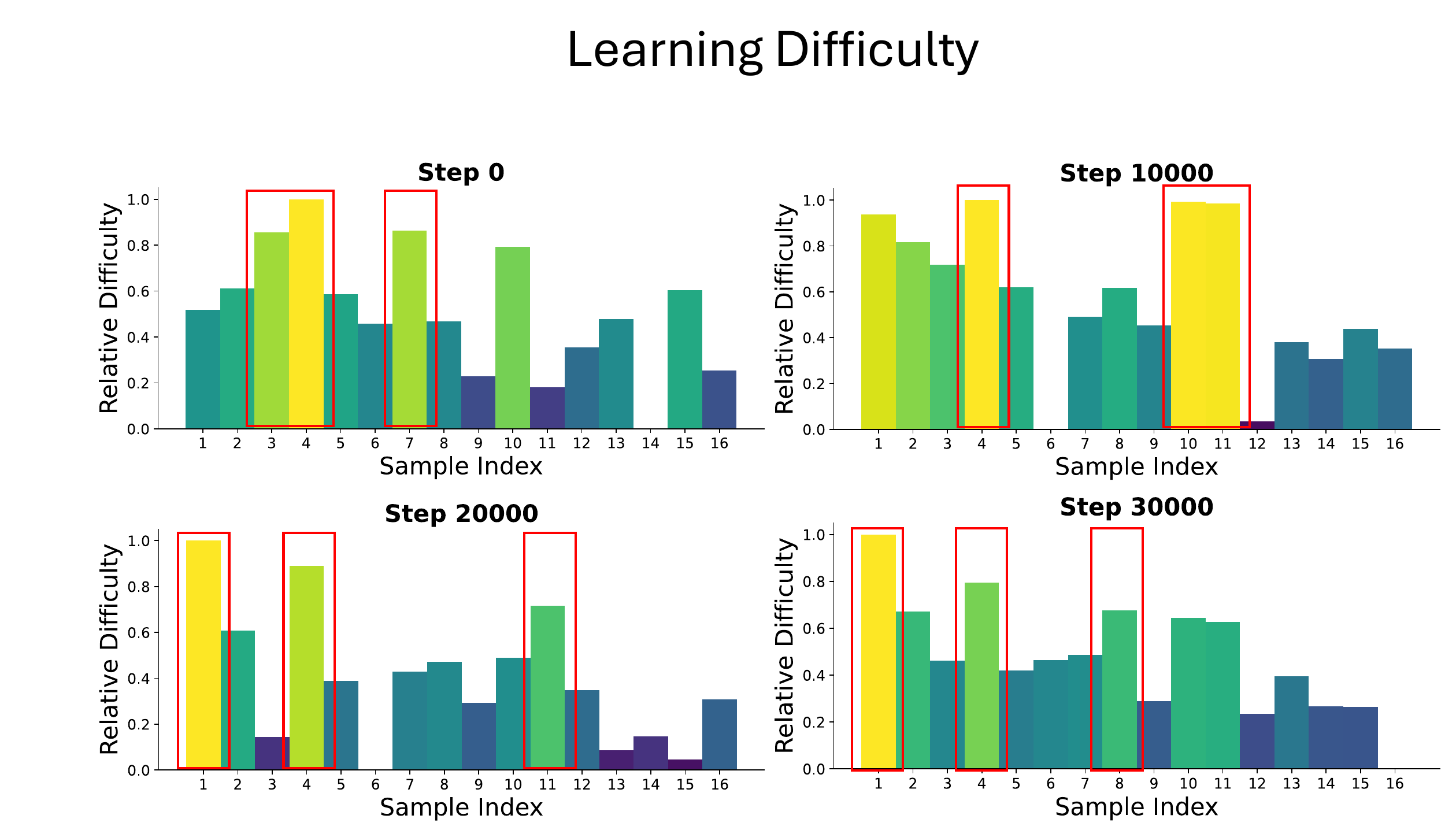}
        \caption{}
        \label{fig:diff}
    \end{subfigure}
    \begin{subfigure}[b]{0.49\textwidth}
        \includegraphics[width=\textwidth]{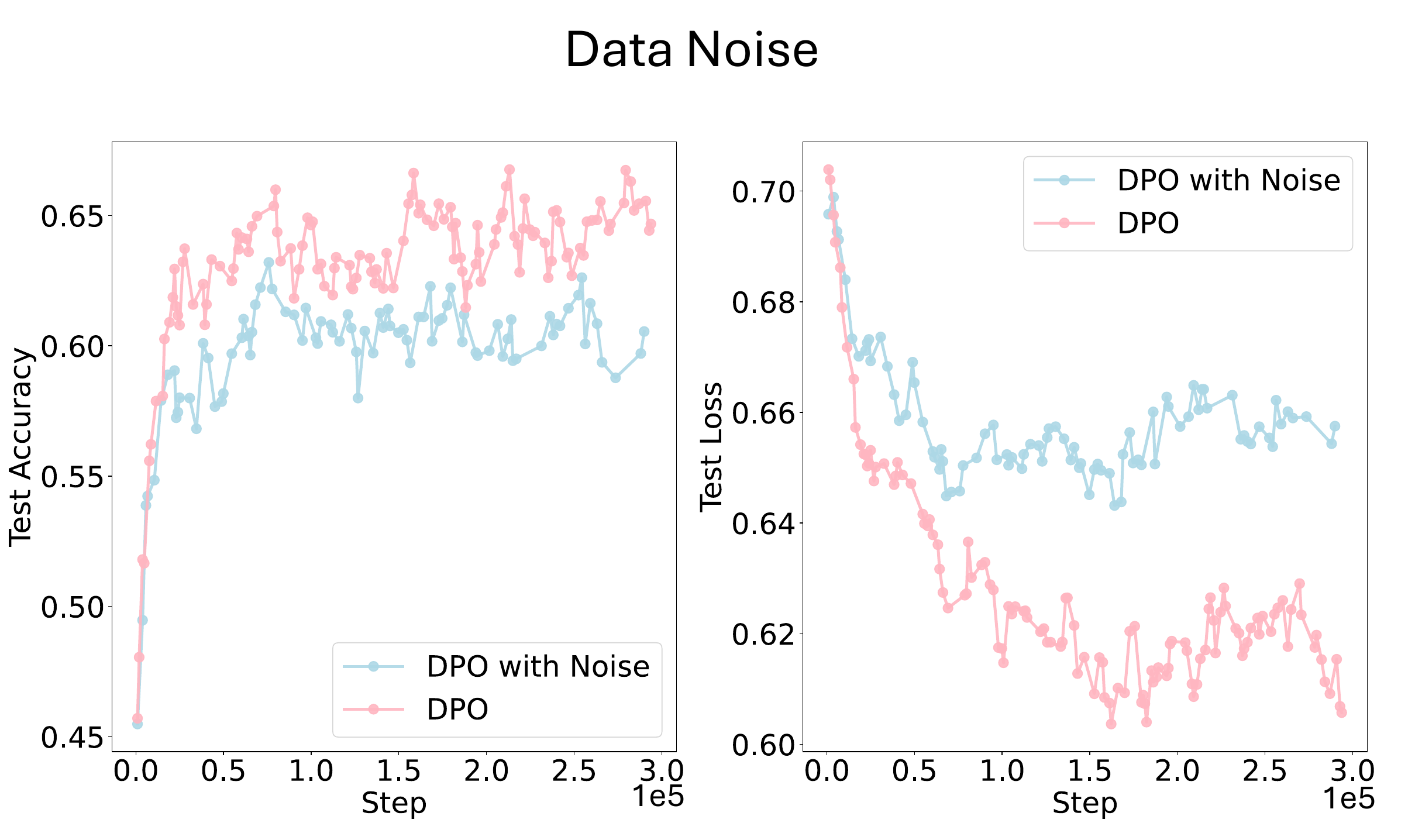}
        \caption{}
        \label{fig:noise}
    \end{subfigure}
        \hfill
     \vspace{-0.5em}
    \caption{The Study of challenges in SamS. (a) \textbf{ Varying learning difficulties for different model states.}
 For the same 16 samples, we track their DPO loss across different states of the language model, from training step 0 to step 30,000.  We use the relative DPO loss of each sample as the difficulty measure~\cite{gao2025principled}. (b) \textbf{Noisy data degrades DPO performance.} 
 During preference optimization using Pythia-2.8B~\cite{biderman2023pythia} on the Anthropic-HH dataset~\cite{bai2022training}, we artificially injected 20\% noise into the preference labels. As a result, the performance of DPO dropped significantly, highlighting its sensitivity to data quality.}
    \label{fig:combined}
\end{figure}

In contrast to these existing studies, this paper introduces a novel problem: Sample Scheduling for DPO.
Specifically, given a fixed preference dataset, the goal is to dynamically and adaptively schedule training samples based on the evolving internal states of the language model during preference optimization. This formulation is motivated by two key challenges: 
First, as shown in Figure~\ref{fig:diff}, samples in the training dataset may exhibit varying levels of learning difficulty for different model states. As the model’s internal state evolves over time, the relative difficulty of each sample may also shift. 
Without an adaptive scheduling mechanism, 
the model may overemphasize samples misaligned with its current learning capacity or overfit to some error patterns, thereby impairing its alignment performance~\cite{gao2025principled,zou2025transformer}.
Second, the dataset may contain noisy samples~\cite{shen2024towards}.  
As shown in Figure~\ref{fig:noise}, incorrect or inconsistent preference labels can destabilize the DPO training process~\cite{gao2024impact}, and low-quality but preferred responses may erode the original conversational ability of the model.
We also empirically verify the presence of such noise in Appendix~\ref{supp:gptjudge}.

To address this problem, we propose a scheduling algorithm \textbf{SamS},  \textbf{Sam}ple \textbf{S}cheduling for Direct Preference Optimization.
In particular, we formulate Sample Scheduling for DPO as a contextual bandit problem, where we define the reward for sample scheduling by leveraging the loss signal during DPO training, and define the arm context based on the internal state representation of LLMs.
In this setting, \sysn employs a scheduler model to adaptively select samples from each training batch according to the model's evolving states, in order to maximize the potential resulting generalization performance.  
It incorporates two key innovations.
First, it adopts a lagged training strategy, where the scheduler is updated in the subsequent training round, allowing the reward to be collected without incurring additional computational overhead. 
Second, it introduces an auxiliary exploration network to explicitly address the exploration-exploitation dilemma that is inherent in the iterative sample scheduling problem.

We conduct extensive experiments across diverse benchmarks, including AlpacaEval 2~\cite{dubois2024length} and MT-Bench~\cite{zheng2023judging}, to evaluate the effectiveness of \sysn. Notably, when integrated with the original DPO loss, \sysn consistently outperforms several advanced offline preference optimization methods on mainstream evaluation benchmarks. Particularly, our method improves the AlpacaEval 2 win rate (WR) by 3.0\% - 12.4\% and the length-controlled win rate (LC) by 5.5\% - 8.4\% compared to the baselines.
Furthermore, we conduct a thorough evaluation of \sysn under noisy preference data conditions and show that its integration significantly enhances robustness against label noise. Importantly, thanks to the carefully designed scheduling reward and the lightweight architecture of \sysn, the added training overhead is minimal, and GPU memory consumption is even reduced.

In summary, our contributions can be summarized as follows:
(1) \textbf{Novel Problem}: We introduce a new problem, Sample Scheduling for DPO, which highlights a promising direction for improving LLM alignment performance using fixed preference datasets. (2) \textbf{Proposed Algorithm}: We propose \sysn, a scheduling algorithm that adaptively selects training samples from each batch according to the model's evolving internal states. (3) \textbf{Empirical Effectiveness}: \sysn can be seamlessly integrated into existing DPO pipelines without modifications to the core algorithm, yielding substantial performance improvements with only marginal additional computational overhead.
Batch-wise sample selection opens a promising path for efficient LLM alignment, and the idea naturally extends to RLHF and other supervised learning paradigms.

\section{Preliminary}

DPO ~\cite{rafailov2023direct} is an offline preference optimization algorithm designed to simplify and stabilize training by reparameterizing the reward function typically used in RLHF.
Specifically, DPO reparameterizes the reward model using a closed-form expression:
\begin{equation}
r(x,y)=\beta \log \frac{\pi_{\theta}(y|x)}{\pi_{\text{ref}}(y|x)} + \beta \log Z(x),
\end{equation}
where $\pi_\theta$ represents the policy model, $\pi_\text{ref}$ is the supervised fine-tuned reference policy, and $Z(x)$ denotes the partition function.

Given a data sample $a = (x,y^w,y^l)$, where $y^w$ and $y^l$ represent the preferred and dispreferred completions respectively for the prompt $x$, the DPO framework incorporates this reward formulation into the Bradley-Terry ranking objective~\cite{bradley1952rank}. Specifically, it defines the probability $p(y^w > y^l| x) = \sigma(r(x, y^w) - r(x, y^l))$, where $\sigma$ denotes the logistic function. Consequently, the objective of DPO is formally defined as:
\begin{equation}
\mathcal{L}_{\text{DPO}}(a; \theta) = -\mathbb{E}_{(x,y^w,y^l) \sim \mathcal{D}}
\left [ \log \sigma \left( \beta \left( \log \frac{\pi_{\theta}(y^w|x)}{\pi_{\text{ref}}(y^w|x)} 
- \log \frac{\pi_{\theta}(y^l|x)}{\pi_{\text{ref}}(y^l|x)} \right) \right) \right].
\end{equation}
In practice, batch-level preference optimization is commonly employed. Given a batch consisting of $n$ samples, denoted as $X_t = \{ a_{t,i} \}_{i=1}^n$, where each sample $a_{t,i} = (x_{t,i}, y^w_{t,i}, y^l_{t,i})$, the average-based DPO loss is formally defined as:
\begin{equation}
\mathcal{L}_{\text{DPO}}(X_t; \theta)=\frac{1}{|X_t|} \sum_{a_{t,i} \in X_t} {\mathcal{L}}_{\text{DPO}}(a_{t,i}; \theta).
\end{equation}
During each training round $t \in [T]$, the policy $\pi_{\theta}$ typically learns from the entire current batch $X_t$, which may contain irrelevant, challenging, or noisy samples. To address this, our objective is to train a scheduler capable of effectively exploring the sample space, thereby identifying and selecting reliable, high-quality samples for the policy's offline preference optimization. 

\section{The Sample Scheduling Problem} \label{framework}

We formulate the Sample Scheduling problem for offline preference optimization using the contextual bandit framework proposed in ~\cite{ban2024neural, ban2021ee, hwang2023combinatorial}. Let $\pi_\theta$ denote a language model parameterized by $\theta$ that we aim to align with human preferences, and let $f$ denote a scheduler designed to perform interactive sample scheduling during batch-level preference optimization.

\paragraph{Problem Formulation.} Assume the learning process spans $T$ rounds. At each round $t \in [T]$, we draw a batch containing $n$ samples, denoted by $X_t = \{a_{t,1}, a_{t,2}, \dots, a_{t,n}\} \sim \mathcal{D}$, where each sample $a_{t,i} = (x_{t,i}, y^w_{t,i}, y^l_{t,i})$ for $i \in [n]$ is considered an arm, resulting in $n$ total arms. For each arm $a_{t,i}$, we define a contextual representation $\bar{x}_{t,i} = h(x_{t,i}, y^w_{t,i}, y^l_{t,i})$, where $h(\cdot)$ is an encoding function mapping each sample to a context representation vector.

Given a subset $\widetilde{X}_t \subset X_t$ with size $K$, $|\widetilde{X}_t| = k$, selected by the scheduler $f$, we train the policy $\pi_{\theta_{t-1}}$ on this subset, updating the policy parameters to $\theta_t$ as follows:
\begin{equation}
   \theta_{t}= \theta_{t-1}-\eta \nabla_{\theta_{t-1}}\mathcal{L}_{\text{DPO}}(\widetilde{X}_t; \theta_{t-1}).
\end{equation}
To measure the improvement from $\theta_{t-1}$ to $\theta_t$ using the selected subset $\widetilde{X}_t$, we introduce a reward function $r(\widetilde{X}_t, \theta_{t-1}\rightarrow \theta_{t})$, which is initially unknown.
In each round $t \in [T]$, the scheduler $f$ selects a subset $\widetilde{X}_t$ from batch $X_t$ and provides it to policy $\pi_{\theta_{t-1}}$. Subsequently, the scheduler observes the reward $r(\widetilde{X}_t, \theta_{t-1}\rightarrow \theta_t)$, which informs updates to its parameters for future scheduling optimization.
The objective for the scheduler $f$ over $T$ rounds is thus to select a sequence of subsets $\{\widetilde{X}_1, \widetilde{X}_2, \dots, \widetilde{X}_T\}$ that maximizes the cumulative reward:
\begin{equation}
\max \sum_{t=1}^{T} r(\widetilde{X}_t, \theta_{t-1} \rightarrow \theta_t).
\end{equation}

\paragraph{Reward Definition.} 
In supervised preference learning, accurately measuring the performance improvement of policy $\pi_{\theta}$ from $\theta_{t-1}$ to $\theta_{t}$ using an oracle is typically impractical. To address this limitation, we propose a reward definition $r$ leveraging insightful information from the learning trajectory of $\theta$. This reward acts as the supervisory signal for training the scheduler $f$ and comprises two distinct components: a batch-level reward and a sample-level reward.

First, we introduce the batch-level reward, which measures the reduction in the average DPO loss before and after training with a selected batch. In practice, we use the batch-average DPO loss to approximate the expected DPO loss across the entire data distribution. Formally, at round $t$, given the policy parameters $\theta_{t-1}$, we train on a subset $\widetilde{X}_t$, resulting in updated parameters $\theta_t$. The batch-level reward for selecting $\widetilde{X}_t$ is defined as:
\begin{equation} \label{eq:batchreward}
r^{B}(X_t, \theta_{t-1}, X_{t+1}, \theta_t) = \frac{\overbrace{\sum_{i=1}^{n} e^{\mathcal{L}_{\text{DPO}}(a_{t, i}; \theta_{t-1})}}^{A} - \overbrace{\sum_{i=1}^{n} e^{\mathcal{L}_{\text{DPO}}(a_{t+1, i}; \theta_{t})}}^{B}}{\max\left(\sum_{i=1}^{n} e^{\mathcal{L}_{\text{DPO}}(a_{t, i}; \theta_{t-1})},\sum_{i=1}^{n} e^{\mathcal{L}_{\text{DPO}}(a_{t+1, i}; \theta_{t})}\right)}.
\end{equation}
Term $A$ evaluates the performance of $\theta_{t-1}$ on batch $X_t$, noting that $\theta_{t-1}$ has not previously encountered $X_t$. Similarly, term $B$ evaluates the performance of $\theta_{t}$ on the new batch $X_{t+1}$, after $\theta_{t}$ has been trained on $\widetilde{X}_t$. To enhance the sensitivity of the reward metric, we exponentiate the DPO loss $e^{\mathcal{L}_{\text{DPO}}(\cdot)}$ and apply normalization through the denominator. Consequently, $r^{B}$ signifies the approximate performance improvement of the policy $\pi$ after the scheduling decision at round $t$.

Next, we introduce a sample-level reward for fine-grained evaluation, complementing the batch-level reward, which only reflects aggregate improvement over $\widetilde{X}_t$. We assign higher rewards to samples with larger preference margins and greater model uncertainty.

Formally, for a data point $a_{t,i} = (x_{t,i}, y^w_{t,i}, y^l_{t,i})$, we define the sample-level reward:
\begin{equation}\label{eq:samplelevel}
\medmath{
r^{S}(a_{t,i}, \theta_{t-1}) = \underbrace{ g\Bigl( \beta\log \frac{\pi_{\theta_{t-1}}(y^w_{t,i}\mid x_{t,i})}{\pi_{\text{ref}}(y^w_{t,i}\mid x_{t,i})} -  \beta \log \frac{\pi_{\theta_{t-1}}(y^l_{t,i}\mid x_{t,i})}{\pi_{\text{ref}}(y^l_{t,i}\mid x_{t,i})} )\Bigr)}_{\text{preference margin}} + \underbrace{\bigl(1 - g(\log\pi_{\theta_{t-1}}(y^w_{t,i}\mid x_{t,i}))\bigr)}_{\text{model uncertainty}}.
}
\end{equation}
The first term rewards samples with larger preference margins under $\theta_{t-1}$, thereby avoiding convergence on ambiguous or noisy examples \cite{meng2024simpo}. The second term promotes selection of high-uncertainty samples, addressing the tendency of policies to produce out-of-distribution outputs during training \cite{lin2024limited,xu2024dpo}. Here, $g(\cdot)$ is a min-max normalization mapping values to $[0,1]$. We provide a more detailed discussion of the design motivations in the Appendix~\ref{sup:reward}.

Finally, we integrate both the batch-level and sample-level rewards to compute the final reward for each individual sample $a_{t,i}$ as follows:
\begin{equation}\label{eq:samplereward}
r(a_{t, i}, \theta_{t-1} \rightarrow \theta_{t} ) = \gamma \sigma\left[r^{B}(X_t, \theta_{t-1}, X_{t+1}, \theta_t) \right] + (1-\gamma)  \sigma \left[r^{S}(a_{t,i}, \theta_{t-1}) \right],
\end{equation}
where $\gamma \in [0,1]$ is a hyperparameter that controls the trade-off between the batch-level and sample-level reward signals, and $\sigma(\cdot)$ denotes the sigmoid function.

To mitigate the combinatorial complexity associated with subset selection, we define the reward of a selected subset $\widetilde{X}_t$ as the sum of the individual sample rewards:
\begin{equation} \label{eq:finalreward}
r(\widetilde{X}_t, \theta_{t-1} \rightarrow \theta_{t}) = \sum_{a_{t,i} \in \widetilde{X}_t} r(a_{t, i}, \theta_{t-1} \rightarrow \theta_{t}).
\end{equation}
In addition to providing valuable insights, the reward can be computed straightforwardly during the DPO process.

\paragraph{Arm Context Design.}
The arm context serves as the input to the scheduling module, with the goal of leveraging the representational capacity of the policy model $\pi_{\theta}$.
For each sample, we extract the intermediate hidden representations from all transformer block layers of $\pi_{\theta}$, and define the arm context as $\bar{x}_{t,i} = h(x_{t,i}, y^w_{t,i}, y^l_{t,i})$.
To obtain a fixed-dimensional vector, we apply a combination of concatenation and pooling operations over the token-level hidden states across layers. This design allows us to take into account the evolution of the state of the LLM into the arm representations.

 \begin{figure}[!t]
  \centering
  \includegraphics[width=1.0\textwidth]{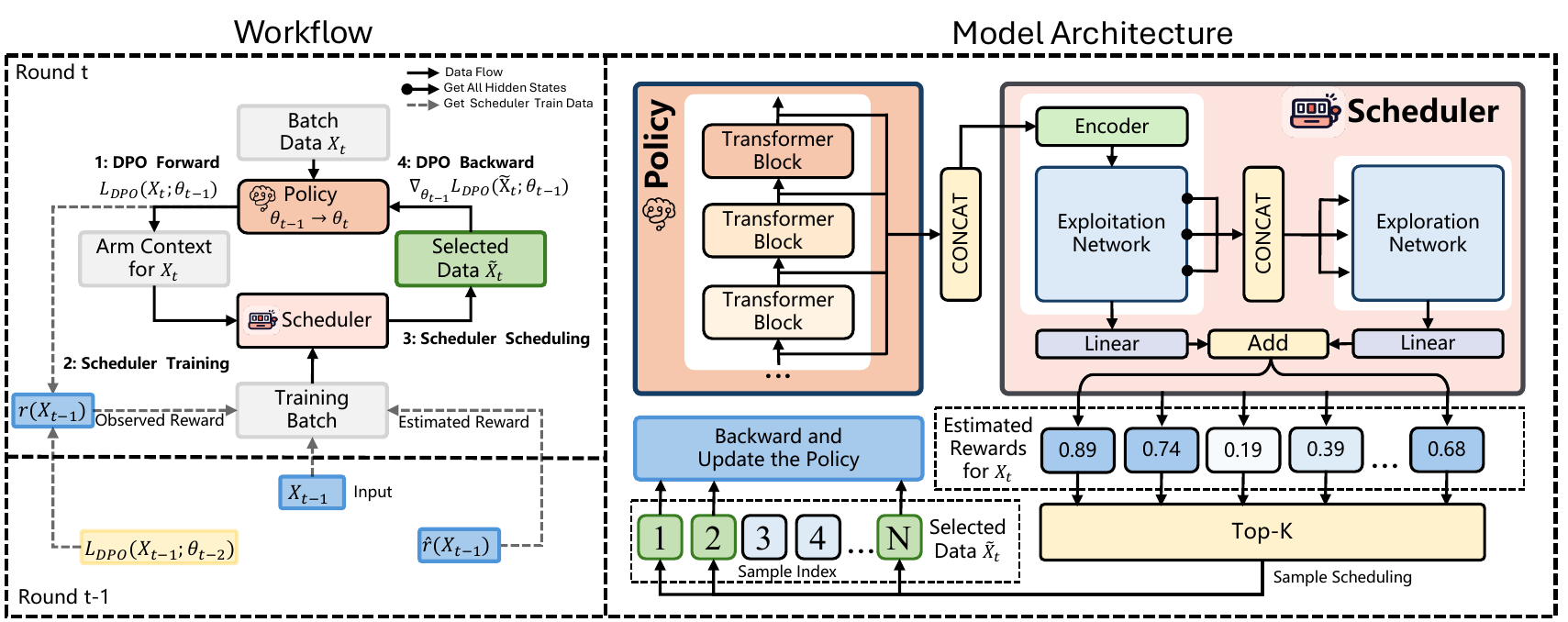}
  \caption{(Left side) Overview of a standard DPO framework integrated with SamS. (Right side) The architecture of the Scheduler. The Scheduler initially treats the policy's hidden state sequence as the arm context for each sample. The Encoder aggregates the state information of each sample to encode the arm context. Subsequently, the Exploitation-Exploration Network utilizes the encoded arm contexts to estimate reward values for each sample, which is used to select a Top-K subset for policy learning.}
  \label{fig:workflow}
\end{figure}

\section{Proposed Algorithm: \sysn} \label{sec:architecture}

We present the overall framework of the proposed method, \textbf{SamS} (\textbf{Sam}ple \textbf{S}cheduling), followed by a detailed description of its model structure and workflow.
Sample Scheduling can be naturally considered as a sequential decision-making problem under uncertainty, where the internal states of the policy $\theta$ evolve across rounds, resulting in uncertainties for the iterative sample selection. Consequently, the exploration-exploitation dilemma is inherently embedded in this problem.

\paragraph{Model Structure.}
As shown in Figure \ref{fig:workflow}, the scheduler $f$ consists of an encoder layer followed by two specialized networks for exploitation and exploration. The Encoder Layer takes the hidden state representations of each sample as input and produces an encoded representation used by the subsequent neural networks. For notational simplicity, we continue to denote the encoded arm context as $\bar{x}_{t,i}$.

Denote the exploitation network by $f^S(\cdot; \theta^S)$ and the exploration network by $f^{S{\prime}}(\cdot; \theta^{S{\prime}})$.
The exploitation network $f^S$ learns to predict the reward of each sample arm by mapping the arm context $\bar{x}_{t,i}$ to its observed reward $r(a_{t,i}, \theta_{t-1} \rightarrow \theta_t)$. 
The exploration network $f^{S{\prime}}$ estimates the uncertainty of the predictions made by $f^S$, and augments the original reward estimate with a potential exploration bonus. This design enables a principled trade-off between exploitation and exploration during iterative sample selection, referring to the design in~\cite{ban2024neural,ban2022improved}.
This process aligns with the principles of classic Exploration-Exploitation algorithms, such as Upper Confidence Bound (UCB)~\cite{ban2021local,ban2024meta,qi2023graph} and Thompson Sampling (TS)~\cite{wen2015efficient,zhang2020neural}.

\begin{algorithm}[!t]
\caption{Proposed Algorithm: \sysn}
\label{alg:practical}
\begin{algorithmic}[1]
\REQUIRE $T$, $n$, $K$, $\theta$ (LLM Parameters), $\theta^S, \theta^{S'}$ (Scheduler Parameters) 
\STATE Initialize $\theta_0, \theta^{S}_1, \theta^{S'}_1$
\FOR{$t = 2, 3, \dots, T$}
    \STATE Draw batch data $X_t \sim \mathcal{D}$     
    \STATEx \colorbox{gray!20}{$\triangledown$ DPO Forward with $X_t$}  
    \STATE Compute DPO Loss $L_{\text{DPO}}(X_t; \theta_{t-1})$    \ \ \# \texttt{\small Standard Forward Pass}
    \STATEx \colorbox{gray!20}{$\triangledown$ Scheduler Training}
    \STATE Compute $r^{B}(X_{t-1}, \theta_{t-2}, X_{t}, \theta_{t-1})$ based Eq.\eqref{eq:batchreward} \ \ \# \texttt{\small Observe Batch-level Reward}
    \FOR{$a_{t-1,i} \in \widetilde{X}_{t-1}$}
    \STATE Compute $r^{S}(a_{t-1, i}, \theta_{t-2})$ according to Eq.\eqref{eq:samplelevel}    \ \ \#  \texttt{\small Observe Sample-level Reward}
    \STATE Compute $ r(a_{t-1, i}, \theta_{t-2} \rightarrow  \theta_{t-1})$ according to Eq.\eqref{eq:samplereward}   \ \ \# \texttt{\small Observe Final Reward}
\ENDFOR 
   \STATE Compute $\mathcal{L}^{S}(\widetilde{X}_{t-1}, \theta_{t-1}^S)$ According to Eq.\eqref{loss1}  
    \STATE $\theta_{t}^S=\theta_{t-1}^S-\eta_1 \nabla_{\theta^S_{t - 1}}\mathcal{L}^{S}(\widetilde{X}_{t-1}, \theta_{t-1}^S)$    \ \ \# \texttt{\small Update Exploitation Network of Scheduler}
    \STATE Compute \ $\mathcal{L}^{S'}(\widetilde{X}_{t-1}, \theta_{t-1}^{S'})$ According to Eq.\eqref{loss2}  
    \STATE $\theta_{t}^{S'}=\theta_{t-1}^{S'}-\eta_2 \nabla_{\theta^{S'}_{t - 1}}\mathcal{L}^{S'}(\widetilde{X}_{t-1}, \theta_{t-1}^{S'})$      \ \ \# \texttt{\small Update Exploration Network of Scheduler}
    \STATEx \colorbox{gray!20}{$\triangledown$ Scheduler Scheduling}
    \FOR{$i \in [n]$}
    \STATE $\hat{r}(a_{t, i}, \theta_{t-1}\rightarrow \theta_{t}) = f^S(\bar{x}_{t,i}; \theta^S_{t}) + \lambda  f^{S'}(h^S_{t,i}; \theta^{S'}_{t})$        \ \ \# \texttt{\small Estimated Reward for Each Sample Based on Exploitation-Exploration Trade-off}
    \ENDFOR
    \STATE $\widetilde{X}_t = \text{Top-$K$}_{i\in [n]} \hat{r}(a_{t, i}, \theta_{t-1}\rightarrow \theta_{t})$ \ \ \# \texttt{\small Choose $\widetilde{X}_t$}
    \STATEx \colorbox{gray!20}{$\triangledown$ DPO Backward with $\widetilde{X}_t$}
    \STATE $\theta_{t}= \theta_{t-1}-\eta \nabla_{\theta_{t-1}}\mathcal{L}_{\text{DPO}}(\widetilde{X}_t; \theta_{t-1})$  \ \ \# \texttt{\small Udpate LLM with $\widetilde{X}_t$}
\ENDFOR
\STATE \textbf{Return:} $\theta_T$
\end{algorithmic} 
\end{algorithm}

Given the input $\bar{x}_{t,i}$ in round $t$, the exploitation network $f^S$ is implemented as a fully connected feedforward neural network with residual connections, denoted by $f^S(\bar{x}_{t,i}; \theta_{t}^S)$.
After receiving the observed reward $r(a_{t,i}, \theta_{t-1} \rightarrow \theta_t)$ in round $t+1$, the parameters $\theta_t^S$ are updated via stochastic gradient descent using the following loss function:
\begin{equation}\label{loss1}
\mathcal{L}^{S}(\widetilde{X}_t, \theta_{t}^S)=\frac{1}{2|\widetilde{X}_t|}\sum_{a_{t,i} \in \widetilde{X}_t}[f^S(\bar{x}_{t,i}; \theta_{t}^S) - r(a_{t,i}, \theta_{t-1} \rightarrow \theta_t)]^2.
\end{equation}
Next, in each round $t \in [T]$, we construct the input to the exploration network $f^{S'}$ by concatenating the intermediate hidden states of $f^S(\bar{x}_{t,i};\theta^S_{t-1})$ along the last dimension, denoted by $h_{t,i}^S$. This design enables the exploration module to take into account the internal states of the exploitation network when making exploration decisions.
The exploration network $f^{S'}$ is also a fully connected feedforward neural network with residual connections.
After receiving the observed reward $r(a_{t,i}, \theta_{t-1} \rightarrow \theta_t)$ in round $t+1$,  the label for training $f^{S'}$ is the difference between observed reward and $f^S(\cdot; \hat{\theta}^S)$ for uncertainty estimation.
The exploration network parameters $\theta_t^{S’}$ are then updated via stochastic gradient descent using the loss:
\begin{equation}\label{loss2}
\mathcal{L}^{S'}(\widetilde{X}_t, \theta_{t}^{S'})=\frac{1}{2|\widetilde{X}_t|}\sum_{a_{t,i} \in \widetilde{X}_t}\left[ f^{S'}(h^S_{t,i}; \theta_{t}^{S'})- \left( r(a_{t,i}, \theta_{t-1} \rightarrow \theta_t)- f^S(\bar{x}_{t,i}; \theta^S_{t}) \right ) \right]^2.
\end{equation}
Finally, the overall reward estimate for each sample is given by: $f(\bar{x}_{t,i}; \theta^S, \theta^{S'}) = f^S(\bar{x}_{t,i}; \theta_{t}^S) + \lambda  f^{S'}(h^S_{t,i}; \theta_{t}^{S'}) $, where $\lambda$ is a tunable hyperparameter controlling the exploration strength.
Next, we describe the training strategy for integrating the scheduler $f$ within the DPO framework.

\paragraph{Workflow.} Algorithm ~\ref{alg:practical} illustrates the workflow of our proposed \sysn algorithm. Each training round consists of four main steps, detailed as follows:

\emph{(1) DPO Forward Pass}. In each training round $t \in [T]$, we first perform a forward pass to compute the DPO loss following the standard DPO procedure (Line 4). We store the loss result of each sample $\mathcal{L}_{\text{DPO}}(a_{t,i}; \theta_{t-1})$ for subsequent scheduler training.

\emph{(2) Scheduler Training}. The objective of this step is to train the scheduler $f$ based on the previously selected subset $\widetilde{X}_{t-1}$ from round $t-1$, utilizing the pair $\{\widetilde{X}_{t-1}, r(\widetilde{X}_{t-1}, \theta_{t-2} \rightarrow \theta_{t-1})\}$. This approach leverages the batch-level reward $r^{B}(X_{t-1}, \theta_{t-2}, X_{t}, \theta_{t-1})$, which requires the loss $\mathcal{L}_{\text{DPO}}(X_{t}; \theta_{t-1})$ computed in the current round $t$, thus avoiding extra computational costs. Lines 5-9 depict the reward calculation for the previously selected subset $\widetilde{X}_{t-1}$, while Lines 10-13 update the scheduler $f$ with the new information. 
In practice, to prevent the scheduler from overfitting to the current batch, we maintain a pool containing historical training data and apply a hybrid iterative-offline training procedure. We display the implementation details in Appendix~\ref{supp:pool}.

\emph{(3) Scheduler Scheduling}. With the updated scheduler parameters $\theta^S_t, \theta^{S'}_{t}$, we estimate rewards for each candidate sample denoted by $\hat{r}(a_{t,i}, \theta_{t-1}\rightarrow \theta_{t})$ as shown in Lines 14-16. Subsequently, we apply a straightforward greedy strategy to select $K$ samples, forming the subset $\widetilde{X}_t$.

\emph{(4) DPO Backward Pass}. Given the selected subset $\widetilde{X}_t$, we compute the corresponding batch loss $\mathcal{L}_{\text{DPO}}(\widetilde{X}_t; \theta_{t-1})$. Since $\widetilde{X}_t$ is a subset of $X_t$, $\mathcal{L}_{\text{DPO}}(\widetilde{X}_t; \theta_{t-1})$ can be efficiently derived from the previously computed $\mathcal{L}_{\text{DPO}}(X_t; \theta_{t-1})$. Finally, the policy model parameters $\theta_{t-1}$ are updated to $\theta_{t}$ through gradient descent (Line 18).

\section{Experiments}

In this section, we present the primary experimental results along with their analysis.
For SamS, both the exploitation and exploration modules are implemented as 16-layer residual MLPs. We set the batch size $|X_t|$ to 64 and the selection size $|\widetilde{X}_t|$ to 32 across all training rounds. Additional implementation details of SamS are provided in Appendix~\ref{detail} due to space constraints.

\subsection{Performance of SamS Embedded in DPO} \label{exp:main}

In this subsection, we evaluate the performance of SamS when integrated into DPO, using widely adopted benchmarks for LLM preference optimization. We compare it against state-of-the-art offline preference optimization methods. Detailed experimental settings can be found in Appendix~\ref{app:expset}.


\begin{table}[ht]
\centering
\caption{AlpacaEval 2~\cite{dubois2024length} and MT-Bench~\cite{zheng2023judging} results under the two model settings. LC and WR denote length-controlled and raw win rate, respectively. Here, \textbf{bold} denotes the best performance, \underline{underline} indicates the second-best performance, and "-" represents that no measurement was taken. }
\vspace{10pt} 

\resizebox{0.8\textwidth}{!}{
\begin{tabular}{lcccccccccc}
\toprule
\multirow{3}{*}{\textbf{Method}} & \multicolumn{3}{c}{\textbf{Mistral-Instruct (7B)} } & \multicolumn{3}{c}{\textbf{Llama3-Instruct (8B)}} \\
\cmidrule(lr){2-4} \cmidrule(lr){5-7}
& \multicolumn{2}{c}{\textbf{AlpacaEval 2}} & \multicolumn{1}{c}{\textbf{MT-Bench}} & \multicolumn{2}{c}{\textbf{AlpacaEval 2}} &  \multicolumn{1}{c}{\textbf{MT-Bench}} \\
\cmidrule(lr){2-3} \cmidrule(lr){4-4} \cmidrule(lr){5-6} \cmidrule(lr){7-7} 
& \textbf{LC (\%)} & \textbf{WR (\%)} & \textbf{ GPT-4 Turbo} & \textbf{LC (\%)} & \textbf{WR (\%)} &  \textbf{GPT-4 Turbo} \\
\midrule
SFT & 17.1  & 14.7  & 6.2 & 26.0  & 25.3  & 6.9 \\
\midrule
RRHF~\cite{yuan2023rrhf} & 25.3 & 24.8  & \underline{6.5} & 31.3 & 28.4  & 6.7 \\
SLiC-HF~\cite{zhao2023slic} & 24.1 & 24.6  & \underline{6.5} & 26.9 & 27.5 & 6.8 \\
IPO~\cite{azar2024general} & 20.3 & 20.3 & 6.4 & 35.6 & 35.6  & \underline{7.0} \\
CPO~\cite{xu2024contrastive} & 23.8 & \underline{28.8} & 6.3 & 28.9 & 32.2  & \underline{7.0} \\
KTO~\cite{ethayarajh2024kto} & 24.5 & 23.6 & 6.4 & 33.1 & 31.8 & 6.9 \\
ORPO~\cite{hong2024orpo} & 24.5 & 24.9 & 6.4 & 28.5 & 27.4  & 6.8 \\
R-DPO~\cite{park2024disentangling} & \underline{27.3} & 24.5  & 6.2  & \underline{41.1} & 37.8  &\underline{7.0} \\
DPO~\cite{rafailov2023direct} & 26.8 & 24.9 & 6.3 & 40.3 & \underline{37.9}  & \underline{7.0} \\
DPO (50\%) & 25.2 & 23.8 & 6.3 & 37.5 & 36.2  & 6.9 \\
\midrule
DPO+SamS & \textbf{33.6} & \textbf{36.2}  & \textbf{6.7} & \textbf{42.2} & \textbf{40.5}  &\textbf{7.1} \\
\midrule
\multirow{3}{*}{\textbf{Method}} & \multicolumn{3}{c}{\textbf{Llama3-Instruct v0.2 (8B)} } & \multicolumn{3}{c}{\textbf{Gemma2-Instruct v0.2 (9B)}} \\
\cmidrule(lr){2-4} \cmidrule(lr){5-7}
& \multicolumn{2}{c}{\textbf{AlpacaEval 2}} & \multicolumn{1}{c}{\textbf{MT-Bench}} & \multicolumn{2}{c}{\textbf{AlpacaEval 2}} &  \multicolumn{1}{c}{\textbf{MT-Bench}} \\
\cmidrule(lr){2-3} \cmidrule(lr){4-4} \cmidrule(lr){5-6} \cmidrule(lr){7-7} 
& \textbf{LC (\%)} & \textbf{WR (\%)} & \textbf{ GPT-4 Turbo} & \textbf{LC (\%)} & \textbf{WR (\%)} &  \textbf{GPT-4 Turbo} \\
\midrule
SFT & 26.0  & 25.3  & 6.9 & 48.14  & 36.5  & - \\
\midrule
RRHF~\cite{yuan2023rrhf} & 37.9 & 31.6  & 7.1 & - & -  & - \\
SLiC-HF~\cite{zhao2023slic} & 33.9 & 32.5  & 6.9 & - & - & - \\
IPO~\cite{azar2024general} & 46.8 & 42.4 & \underline{7.2} & 62.6 & 58.4  & - \\
CPO~\cite{xu2024contrastive} & 34.1 & 36.4& \underline{7.2} & 56.4 & 53.4  & - \\
KTO~\cite{ethayarajh2024kto} & 34.1 & 32.1 & \underline{7.2} & 61.7 & 55.5 & - \\
ORPO~\cite{hong2024orpo} & 38.1 & 33.8 & \underline{7.2} & 56.2 & 46.7  & - \\
R-DPO~\cite{park2024disentangling} & 48.0 & 45.8  & 7.0  & 68.3 & \underline{66.9} & - \\
DPO~\cite{rafailov2023direct} & \underline{48.2} & \underline{47.5} & 7.0 & \underline{70.4} & \underline{66.9}  & - \\
DPO (50\%) & 46.0 & 45.2 & 6.9 & 66.1 & 63.5 & - \\
\midrule
DPO+SamS & \textbf{51.5} & \textbf{48.2} & \textbf{7.3} & \textbf{70.8} & \textbf{67.1} & - \\
\bottomrule
\end{tabular}
}
\label{tab:main}
\end{table}

\textbf{(1) DPO+SamS consistently achieves superior performance.} As shown in Table~\ref{tab:main}, the adaptive sample scheduling mechanism of SamS enables DPO to attain the highest scores across all evaluation metrics. Specifically, DPO+SamS outperforms the best-performing baseline by margins ranging from 0.4\% to 6.3\% on the AlpacaEval 2 LC win rate, from 0.2\% to 7.4\% on the AlpacaEval 2 win rate, and by 0.1 to 0.2  on the MT-Bench score across various settings. These results underscore the broad applicability of SamS in preference optimization and its effectiveness in aligning large language models with human preferences.

\textbf{(2) SamS reliably prioritizes samples that are well-suited to the current model state.} To highlight the sample quality, we compare DPO+SamS against a baseline variant denoted as DPO (50\%), in which 50\% of the training samples in each batch are randomly selected under the same conditions. Across all model configurations, DPO+SamS consistently improves performance over DPO (50\%), with gains of 5.5\% - 8.4\% on the AlpacaEval 2 LC win rate, 3.0\% - 12.4\% on the AlpacaEval 2 win rate, and 0.2 - 0.4 on the MT-Bench score. These substantial improvements demonstrate the effectiveness of SamS in dynamically identifying and utilizing high-quality training samples.

\subsection{Generalization Ability}\label{exp:epoch}

To assess the generalization ability of SamS, we apply SamS to various offline preference optimization algorithms, conducting multi-epoch experiments under diverse preference datasets.

We utilize the pretrained Pythia-2.8B~\cite{biderman2023pythia} as the policy model, using Anthropic-HH~\cite{bai2022training} and SHP~\cite{ethayarajh2110understanding} as the preference dataset. Initially, we perform SFT using the prompts and chosen responses from the  dataset.  
Subsequently, we apply SamS to DPO and KTO, conducting multi-epoch training until test accuracy converges. We then compare the performance metrics of our approach with those of the original methods. For the DPO and KTO loss, we set $\beta = 0.1$.

\begin{table}[ht]
\centering
\caption{Performance improvements (in test accuracy) achieved by integrating SamS with different preference optimization methods.
}
\vspace{10pt} 
\resizebox{0.7\textwidth}{!}{
\begin{tabular}{cccccc}
\toprule
\textbf{Dataset} & \textbf{Method} & \textbf{Test-Acc($\%$)} & \textbf{Dataset} & \textbf{Method} & \textbf{Test-Acc($\%$)}\\
\midrule 
\multirow{6}{*}{\textbf{HH}} 
& \textbf{DPO} & 64.3 & \multirow{6}{*}{\textbf{SHP}} & \textbf{DPO} & 67.6\\
& \textbf{DPO+SamS} & 67.1 &  & \textbf{DPO+SamS} & 70.0\\
& \textbf{Improvement} & \bf +2.8 &  & \textbf{Improvement} & \bf +2.4\\
\cmidrule{2-3}\cmidrule{5-6}
& \textbf{KTO} & 60.2 &  & \textbf{KTO} & 65.2\\
& \textbf{KTO+SamS} & 63.3 &  & \textbf{KTO+SamS} & 67.5\\
& \textbf{Improvement} & \bf +3.1 &  & \textbf{Improvement} & \bf +2.3\\
\bottomrule 
\end{tabular}
}
\label{tab:exp1}
\end{table}

\textbf{(1) Integrating SamS with different offline preference optimization methods consistently enhances performance.}
As shown in Table~\ref{tab:exp1} (with detailed results in Table~\ref{tab:detail_exp1}), applying SamS to two baseline methods yields notable improvements: an average increase of 2.65\% in test accuracy (Test-Acc), a 19.9\% improvement in the reward value of the preferred response (Chosen Reward), and a 5.8\% gain in the log-probability of the preferred response (Chosen Logps). Remarkably, these performance gains are achieved with only 50\% of the original training data, highlighting the sample efficiency of SamS. These results demonstrate that SamS significantly improves both the effectiveness and efficiency of training by prioritizing high-quality samples.

\textbf{(2) SamS effectively mitigates out-of-distribution (OOD) challenges for difficult samples.}
The observed improvements in Chosen Reward and Chosen Logps suggest that the policy's implicit reward model is better optimized, enabling it to assign higher rewards to preferred but hard responses. This outcome aligns with the motivation presented in Section~\ref{framework}, confirming that SamS successfully addresses OOD issues by adaptively focusing on the most informative training samples.

\subsection{SamS Enhances the Robustness of DPO to Label Noise}

\begin{wrapfigure}[10]{r}{0.4\textwidth}
  \centering
  \vspace{-18pt}
  \includegraphics[width=0.4\textwidth]{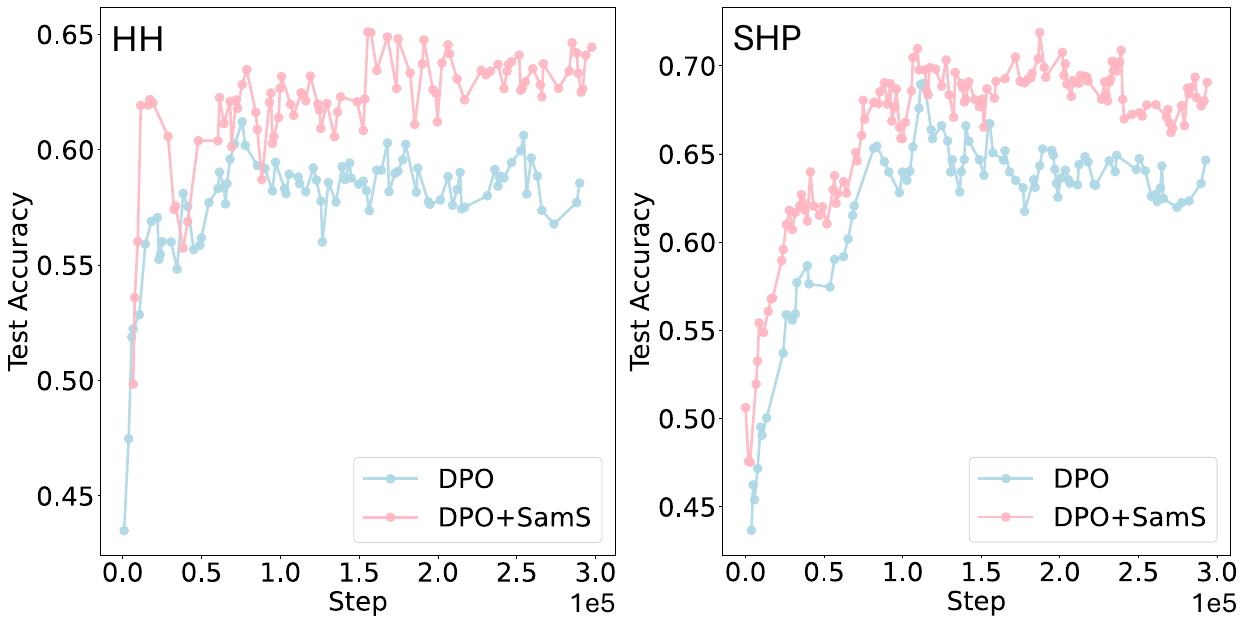}
  \caption{Robustness Testing of SamS: DPO vs. DPO+SamS  (Test Accuracy).}
  \label{fig:robust}
\end{wrapfigure}

To further validate SamS's reliability in selecting high-quality samples from another perspective, we construct a scenario with a contaminated dataset, focusing on its capability to prevent noisy samples from disrupting policy training. Concretely, we randomly flip the preference labels for 20\% of the response pairs in the Anthropic-HH dataset (SHP dataset) and run DPO and DPO+SamS on this modified dataset, adopting the same experimental setup as described in Section~\ref{exp:epoch}.

As illustrated in Figure~\ref{fig:robust}, under the influence of noisy samples, DPO's test accuracy in the HH (resp.~SHP) dataset converges to approximately 58\% (resp.~64\%), a 6\% (resp.~4\%) decline compared to 64\% (resp.~68\%) in the noise-free setting.
In contrast, DPO+SamS converges to around 64\% (68\%), with only a 3\% (2\%) drop from its original 67\% (70\%). DPO+SamS consistently and stably outperforms DPO by approximately 6\% (4\%) in test accuracy, demonstrating superior performance in noisy conditions. Moreover, when compared to the original Anthropic-HH (SHP) dataset, DPO+SamS shows only marginal performance degradation, indicating that \textbf{SamS can effectively maintain the stability of policy training in noisy scenarios}. This is especially crucial in offline preference optimization, where high-quality, manually annotated preference datasets are limited.

\begin{wrapfigure}[12]{r}{0.4 \textwidth}
    \vspace{-5mm}
  \centering
  \includegraphics[width=0.4\textwidth]{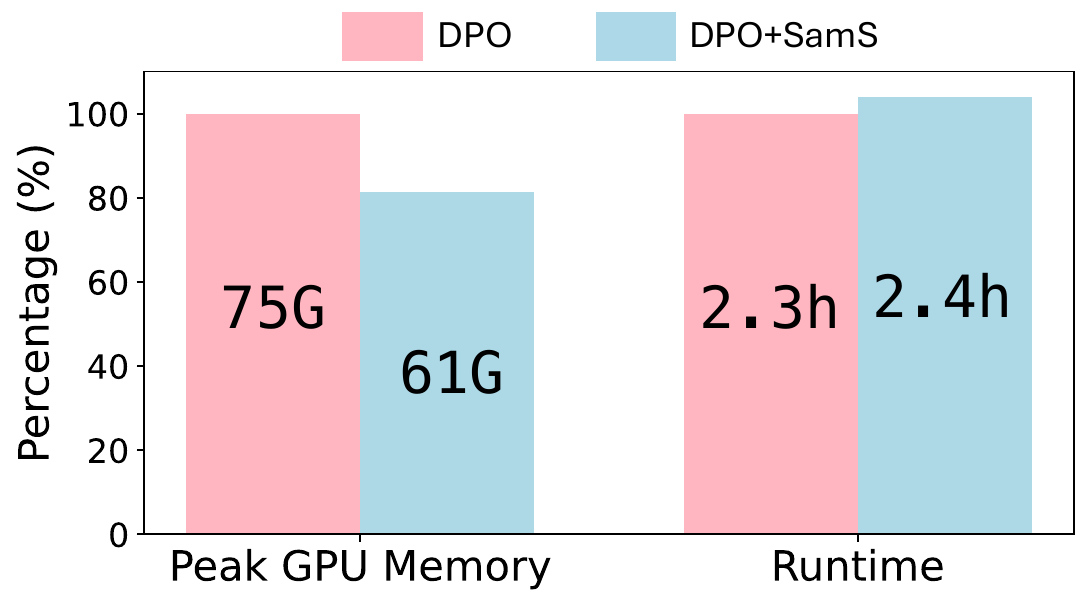}
  \caption{Computational cost  of DPO vs. DPO+SamS: similar runtime and 18\% less GPU memory usage.}
  \label{fig:usage}
\end{wrapfigure}

\subsection{Computational Cost Analysis} \label{compute}
SamS is lightweight and compute-efficient.
Figure~\ref{fig:usage} illustrates the peak single-GPU memory usage and overall runtime of DPO and DPO+SamS under setting of LLaMA environment. Compared to the vanilla DPO implementation, DPO+SamS reduces GPU memory usage by approximately 18\% with similar runtime, owing to SamS's reduction in computational overhead (fewer samples) during backward propagation of LLM updates.
As the reward computation in \sysn does not require additional forward passes through the LLM, and the scheduler model is relatively lightweight, the additional computational cost (running time) introduced by \sysn is marginal.

\subsection{Comparison with Data Pre-Selection} \label{selective dpo}

In this section, we compare SamS with Selective DPO~\cite{gao2025principled}, a representative method of the Data Pre-Selection~\cite{shen2024towards, deng2025less, gao2025principled}, which is most relevant to our problem setting. Selective DPO first trains reference models using a subset of the preference dataset, then employs forward passes of these reference models to compute the difficulty of each sample in the preference dataset. Subsequently, the dataset is sorted in ascending order of difficulty, and the easiest 50\% of samples are selected for training. 

We conduct experiments using the first LLaMA setting, evaluating both Selective DPO and Selective DPO+SamS. In the latter, we further apply SamS to select 75\% of samples in each batch for policy learning based on the ordered subset chosen by Selective DPO.

\textbf{(1) SamS achieves performance comparable to Selective DPO while introducing minimal additional computational cost.} As shown in Table~\ref{tab:selective_dpo}, DPO+SamS yields results similar to those of Selective DPO. However, unlike Selective DPO, which requires a complete additional training phase, SamS can be seamlessly integrated into DPO, incurring only marginal computational overhead. Specifically, Selective DPO entails a total computation time of 6.0 hours, including 5.1 hours for training reference models and 1.2 hours for DPO training. In contrast, our method requires approximately 2.4 hours in total, closely aligning with the time cost of standard DPO while reducing GPU usage by 18\%.

\textbf{(2) Selective DPO+SamS achieves significant performance improvements.} As shown in Table~\ref{tab:selective_dpo}, while both DPO+SamS and Selective DPO effectively enhance performance over the SFT model, Selective DPO+SamS significantly outperforms them. Specifically, Selective DPO+SamS achieves a 46.5\% AlpacaEval 2 LC win rate, a 44.0\% AlpacaEval 2 win rate, and a MT-Bench score of 7.2, representing improvements of 6.2\%, 6.1\%, and 0.2 respectively over DPO.
These significant performance improvements strongly demonstrate the enormous potential of our adaptive sample scheduling strategy when integrated with Data Pre-selection methods.

\begin{table}[ht]
\centering

\caption{The  comparative results of SamS applied on DPO and Selective DPO under the first LLaMA setting. Here, \textbf{bold} denotes the best performance, \underline{underline} indicates the second-best performance, and "-" represents that no measurement was taken.}
\vspace{10pt} 
\resizebox{0.7\textwidth}{!}{
\begin{tabular}{lcccc}
\toprule
\multirow{3}{*}{\textbf{Method}} & \multicolumn{2}{c}{\textbf{AlpacaEval 2}} & \multicolumn{1}{c}{\textbf{MT-Bench}} &\multirow{3}{*}{\textbf{Runtime}}\\
\cmidrule(lr){2-3} \cmidrule(lr){4-4} 
& \textbf{LC (\%)} & \textbf{WR (\%)} & \textbf{GPT-4 Turbo} \\
\midrule
SFT & 26.0  & 25.3  & 6.9 & - \\
\midrule
DPO~\cite{rafailov2023direct} & 40.3 & 37.9  & 7.0 & 2.3 h\\
DPO+SamS & \underline{42.2} &40.5  & \underline{7.1} & 2.4 h\\
Selective DPO~\cite{gao2025principled} &41.7 & \underline{40.9}  & 7.0 & 6.0+1.2 h \\
Selective DPO+SamS &\textbf{46.5} & \textbf{44.0}  &\textbf{7.2} & 6.0+1.3 h \\

\bottomrule
\end{tabular}
}
\label{tab:selective_dpo}
\end{table}

\textbf{For the ablation study, refer to Appendix \ref{app:abstudy}. }

\section{Related Work}

\paragraph{Direct Preference Optimization Variants.} A variety of offline preference optimization algorithms have been proposed besides DPO. Ranking objectives allow for comparisons among more than two instances~\cite{dong2023raft, liu2024lipo, song2024preference, yuan2023rrhf}. Another line of work explores simpler preference optimization objectives that do not rely on a reference model~\cite{Hong2024ORPOMP, xu2023some}.~\cite{zheng2024weak} focuses on post-training extrapolation between the SFT and the aligned model to further enhance model performance.~\cite{bansal2024comparing} proposes a method to jointly optimize instructions and responses, finding it effectively improves DPO.  In this work, we compare DPO+SamS to a series of offline algorithms, including RRHF~\cite{yuan2023rrhf}, SLiC-HF~\cite{Zhao2023SLiCHFSL}, DPO~\cite{Rafailov2023DirectPO}, IPO~\cite{Azar2023AGT}, CPO~\cite{xu2024contrastive}, KTO~\cite{Ethayarajh2024KTOMA}, ORPO~\cite{Hong2024ORPOMP}, and R-DPO~\cite{Park2024DisentanglingLF}, and find that DPO+SamS can outperform them while achieving remarkably high sample efficiency.

\paragraph{Iterative Direct Preference Optimization .} 
The absence of an explicit reward model in DPO limits its capability to sample preference pairs from the optimal policy.~\cite{dong2024rlhf, Kim2024sDPODU, Rosset2024DirectNO, xiong2024iterative,yuan2024self} extend the preference data augmentation approach~\cite{Zhao2023SLiCHFSL,liu2024statistical,he2024llm} to an iterative training framework, where the reference model is continuously updated with the latest policy model or new preference pairs are generated at each iteration.
In this study, we concentrate solely on offline settings.

\section{Conclusion}

We introduce a novel problem setting, Sample Scheduling for DPO, which highlights a promising direction for enhancing  LLM alignment performance using fixed preference datasets. To address this problem, we propose SamS, an efficient adaptive algorithm that dynamically selects training samples from each batch based on the model's evolving state. Without modifying the underlying DPO algorithm, simply integrating SamS into the framework achieves significant performance improvements while incurring only marginal additional computational costs.

\newpage
\section*{Acknowledgments}
This work was supported by the National Natural Science Foundation of China (No. 62276015 and No. 62506024).

\bibliographystyle{plain}
\bibliography{neurips_2025}

@article{christiano2017deep,
  title={Deep reinforcement learning from human preferences},
  author={Christiano, Paul F and Leike, Jan and Brown, Tom and Martic, Miljan and Legg, Shane and Amodei, Dario},
  journal={Advances in neural information processing systems},
  volume={30},
  year={2017}
}

@article{ouyang2022training,
  title={Training language models to follow instructions with human feedback},
  author={Ouyang, Long and Wu, Jeffrey and Jiang, Xu and Almeida, Diogo and Wainwright, Carroll and Mishkin, Pamela and Zhang, Chong and Agarwal, Sandhini and Slama, Katarina and Ray, Alex and others},
  journal={Advances in neural information processing systems},
  volume={35},
  pages={27730--27744},
  year={2022}
}

@article{glaese2022improving,
  title={Improving alignment of dialogue agents via targeted human judgements},
  author={Glaese, Amelia and McAleese, Nat and Tr{\k{e}}bacz, Maja and Aslanides, John and Firoiu, Vlad and Ewalds, Timo and Rauh, Maribeth and Weidinger, Laura and Chadwick, Martin and Thacker, Phoebe and others},
  journal={arXiv preprint arXiv:2209.14375},
  year={2022}
}

@article{chakraborty2024maxmin,
  title={MaxMin-RLHF: Alignment with diverse human preferences},
  author={Chakraborty, Souradip and Qiu, Jiahao and Yuan, Hui and Koppel, Alec and Huang, Furong and Manocha, Dinesh and Bedi, Amrit Singh and Wang, Mengdi},
  journal={arXiv preprint arXiv:2402.08925},
  year={2024}
}

@article{schulman2017proximal,
  title={Proximal policy optimization algorithms},
  author={Schulman, John and Wolski, Filip and Dhariwal, Prafulla and Radford, Alec and Klimov, Oleg},
  journal={arXiv preprint arXiv:1707.06347},
  year={2017}
}

@article{zheng2023judging,
  title={Judging llm-as-a-judge with mt-bench and chatbot arena},
  author={Zheng, Lianmin and Chiang, Wei-Lin and Sheng, Ying and Zhuang, Siyuan and Wu, Zhanghao and Zhuang, Yonghao and Lin, Zi and Li, Zhuohan and Li, Dacheng and Xing, Eric and others},
  journal={Advances in Neural Information Processing Systems},
  volume={36},
  pages={46595--46623},
  year={2023}
}

@article{rafailov2023direct,
  title={Direct preference optimization: Your language model is secretly a reward model},
  author={Rafailov, Rafael and Sharma, Archit and Mitchell, Eric and Manning, Christopher D and Ermon, Stefano and Finn, Chelsea},
  journal={Advances in Neural Information Processing Systems},
  volume={36},
  pages={53728--53741},
  year={2023}
}

@article{ji2024reinforcement,
  title={Reinforcement learning from human feedback with active queries},
  author={Ji, Kaixuan and He, Jiafan and Gu, Quanquan},
  journal={arXiv preprint arXiv:2402.09401},
  year={2024}
}

@article{muldrew2024active,
  title={Active preference learning for large language models},
  author={Muldrew, William and Hayes, Peter and Zhang, Mingtian and Barber, David},
  journal={arXiv preprint arXiv:2402.08114},
  year={2024}
}

@article{liu2024sample,
  title={Sample-efficient alignment for llms},
  author={Liu, Zichen and Chen, Changyu and Du, Chao and Lee, Wee Sun and Lin, Min},
  journal={arXiv preprint arXiv:2411.01493},
  year={2024}
}

@inproceedings{hwang2023combinatorial,
  title={Combinatorial neural bandits},
  author={Hwang, Taehyun and Chai, Kyuwook and Oh, Min-hwan},
  booktitle={International Conference on Machine Learning},
  pages={14203--14236},
  year={2023},
  organization={PMLR}
}

@inproceedings{dudik2015contextual,
  title={Contextual dueling bandits},
  author={Dud{\'\i}k, Miroslav and Hofmann, Katja and Schapire, Robert E and Slivkins, Aleksandrs and Zoghi, Masrour},
  booktitle={Conference on Learning Theory},
  pages={563--587},
  year={2015},
  organization={PMLR}
}

@article{ban2021ee,
  title={Ee-net: Exploitation-exploration neural networks in contextual bandits},
  author={Ban, Yikun and Yan, Yuchen and Banerjee, Arindam and He, Jingrui},
  journal={arXiv preprint arXiv:2110.03177},
  year={2021}
}

@article{ban2024neural,
  title={Neural active learning beyond bandits},
  author={Ban, Yikun and Agarwal, Ishika and Wu, Ziwei and Zhu, Yada and Weldemariam, Kommy and Tong, Hanghang and He, Jingrui},
  journal={arXiv preprint arXiv:2404.12522},
  year={2024}
}

@inproceedings{wen2015efficient,
  title={Efficient learning in large-scale combinatorial semi-bandits},
  author={Wen, Zheng and Kveton, Branislav and Ashkan, Azin},
  booktitle={International Conference on Machine Learning},
  pages={1113--1122},
  year={2015},
  organization={PMLR}
}

@article{zhang2020neural,
  title={Neural thompson sampling},
  author={Zhang, Weitong and Zhou, Dongruo and Li, Lihong and Gu, Quanquan},
  journal={arXiv preprint arXiv:2010.00827},
  year={2020}
}

@article{bradley1952rank,
  title={Rank analysis of incomplete block designs: I. The method of paired comparisons},
  author={Bradley, Ralph Allan and Terry, Milton E},
  journal={Biometrika},
  volume={39},
  number={3/4},
  pages={324--345},
  year={1952},
  publisher={JSTOR}
}

@article{shang2024llava,
  title={Llava-prumerge: Adaptive token reduction for efficient large multimodal models},
  author={Shang, Yuzhang and Cai, Mu and Xu, Bingxin and Lee, Yong Jae and Yan, Yan},
  journal={arXiv preprint arXiv:2403.15388},
  year={2024}
}

@inproceedings{arif2025hired,
  title={HiRED: Attention-Guided Token Dropping for Efficient Inference of High-Resolution Vision-Language Models},
  author={Arif, Kazi Hasan Ibn and Yoon, JinYi and Nikolopoulos, Dimitrios S and Vandierendonck, Hans and John, Deepu and Ji, Bo},
  booktitle={Proceedings of the AAAI Conference on Artificial Intelligence},
  volume={39},
  number={2},
  pages={1773--1781},
  year={2025}
}

@article{vaswani2017attention,
  title={Attention is all you need},
  author={Vaswani, Ashish and Shazeer, Noam and Parmar, Niki and Uszkoreit, Jakob and Jones, Llion and Gomez, Aidan N and Kaiser, {\L}ukasz and Polosukhin, Illia},
  journal={Advances in neural information processing systems},
  volume={30},
  year={2017}
}

@article{liu2023visual,
  title={Visual instruction tuning},
  author={Liu, Haotian and Li, Chunyuan and Wu, Qingyang and Lee, Yong Jae},
  journal={Advances in neural information processing systems},
  volume={36},
  pages={34892--34916},
  year={2023}
}

@article{meng2024simpo,
  title={Simpo: Simple preference optimization with a reference-free reward},
  author={Meng, Yu and Xia, Mengzhou and Chen, Danqi},
  journal={Advances in Neural Information Processing Systems},
  volume={37},
  pages={124198--124235},
  year={2024}
}

@article{lin2024limited,
  title={On the limited generalization capability of the implicit reward model induced by direct preference optimization},
  author={Lin, Yong and Seto, Skyler and Ter Hoeve, Maartje and Metcalf, Katherine and Theobald, Barry-John and Wang, Xuan and Zhang, Yizhe and Huang, Chen and Zhang, Tong},
  journal={arXiv preprint arXiv:2409.03650},
  year={2024}
}

@article{xu2024dpo,
  title={Is dpo superior to ppo for llm alignment? a comprehensive study},
  author={Xu, Shusheng and Fu, Wei and Gao, Jiaxuan and Ye, Wenjie and Liu, Weilin and Mei, Zhiyu and Wang, Guangju and Yu, Chao and Wu, Yi},
  journal={arXiv preprint arXiv:2404.10719},
  year={2024}
}

@article{yuan2023rrhf,
  title={Rrhf: Rank responses to align language models with human feedback without tears},
  author={Yuan, Zheng and Yuan, Hongyi and Tan, Chuanqi and Wang, Wei and Huang, Songfang and Huang, Fei},
  journal={arXiv preprint arXiv:2304.05302},
  year={2023}
}

@article{zhao2023slic,
  title={Slic-hf: Sequence likelihood calibration with human feedback},
  author={Zhao, Yao and Joshi, Rishabh and Liu, Tianqi and Khalman, Misha and Saleh, Mohammad and Liu, Peter J},
  journal={arXiv preprint arXiv:2305.10425},
  year={2023}
}

@inproceedings{azar2024general,
  title={A general theoretical paradigm to understand learning from human preferences},
  author={Azar, Mohammad Gheshlaghi and Guo, Zhaohan Daniel and Piot, Bilal and Munos, Remi and Rowland, Mark and Valko, Michal and Calandriello, Daniele},
  booktitle={International Conference on Artificial Intelligence and Statistics},
  pages={4447--4455},
  year={2024},
  organization={PMLR}
}

@article{xu2024contrastive,
  title={Contrastive preference optimization: Pushing the boundaries of llm performance in machine translation},
  author={Xu, Haoran and Sharaf, Amr and Chen, Yunmo and Tan, Weiting and Shen, Lingfeng and Van Durme, Benjamin and Murray, Kenton and Kim, Young Jin},
  journal={arXiv preprint arXiv:2401.08417},
  year={2024}
}

@article{dubois2024length,
  title={Length-controlled alpacaeval: A simple way to debias automatic evaluators},
  author={Dubois, Yann and Galambosi, Bal{\'a}zs and Liang, Percy and Hashimoto, Tatsunori B},
  journal={arXiv preprint arXiv:2404.04475},
  year={2024}
}

@article{ethayarajh2024kto,
  title={Kto: Model alignment as prospect theoretic optimization},
  author={Ethayarajh, Kawin and Xu, Winnie and Muennighoff, Niklas and Jurafsky, Dan and Kiela, Douwe},
  journal={arXiv preprint arXiv:2402.01306},
  year={2024}
}

@article{hong2024orpo,
  title={Orpo: Monolithic preference optimization without reference model},
  author={Hong, Jiwoo and Lee, Noah and Thorne, James},
  journal={arXiv preprint arXiv:2403.07691},
  year={2024}
}

@article{park2024disentangling,
  title={Disentangling length from quality in direct preference optimization},
  author={Park, Ryan and Rafailov, Rafael and Ermon, Stefano and Finn, Chelsea},
  journal={arXiv preprint arXiv:2403.19159},
  year={2024}
}

@inproceedings{biderman2023pythia,
  title={Pythia: A suite for analyzing large language models across training and scaling},
  author={Biderman, Stella and Schoelkopf, Hailey and Anthony, Quentin Gregory and Bradley, Herbie and O’Brien, Kyle and Hallahan, Eric and Khan, Mohammad Aflah and Purohit, Shivanshu and Prashanth, USVSN Sai and Raff, Edward and others},
  booktitle={International Conference on Machine Learning},
  pages={2397--2430},
  year={2023},
  organization={PMLR}
}

@article{bai2022training,
  title={Training a helpful and harmless assistant with reinforcement learning from human feedback},
  author={Bai, Yuntao and Jones, Andy and Ndousse, Kamal and Askell, Amanda and Chen, Anna and DasSarma, Nova and Drain, Dawn and Fort, Stanislav and Ganguli, Deep and Henighan, Tom and others},
  journal={arXiv preprint arXiv:2204.05862},
  year={2022}
}

@article{jiang2023llm,
  title={Llm-blender: Ensembling large language models with pairwise ranking and generative fusion},
  author={Jiang, Dongfu and Ren, Xiang and Lin, Bill Yuchen},
  journal={arXiv preprint arXiv:2306.02561},
  year={2023}
}

@article{cui2023ultrafeedback,
  title={Ultrafeedback: Boosting language models with high-quality feedback},
  author={Cui, Ganqu and Yuan, Lifan and Ding, Ning and Yao, Guanming and Zhu, Wei and Ni, Yuan and Xie, Guotong and Liu, Zhiyuan and Sun, Maosong},
  year={2023}
}

@article{deng2025less,
  title={Less is More: Improving LLM Alignment via Preference Data Selection},
  author={Deng, Xun and Zhong, Han and Ai, Rui and Feng, Fuli and Wang, Zheng and He, Xiangnan},
  journal={arXiv preprint arXiv:2502.14560},
  year={2025}
}

@article{shen2024towards,
  title={Towards data-centric rlhf: Simple metrics for preference dataset comparison},
  author={Shen, Judy Hanwen and Sharma, Archit and Qin, Jun},
  journal={arXiv preprint arXiv:2409.09603},
  year={2024}
}

@article{gao2025principled,
  title={Principled data selection for alignment: The hidden risks of difficult examples},
  author={Gao, Chengqian and Li, Haonan and Liu, Liu and Xie, Zeke and Zhao, Peilin and Xu, Zhiqiang},
  journal={arXiv preprint arXiv:2502.09650},
  year={2025}
}

@article{xia2024less,
  title={Less: Selecting influential data for targeted instruction tuning},
  author={Xia, Mengzhou and Malladi, Sadhika and Gururangan, Suchin and Arora, Sanjeev and Chen, Danqi},
  journal={arXiv preprint arXiv:2402.04333},
  year={2024}
}

@article{xiong2023iterative,
  title={Iterative preference learning from human feedback: Bridging theory and practice for rlhf under kl-constraint},
  author={Xiong, Wei and Dong, Hanze and Ye, Chenlu and Wang, Ziqi and Zhong, Han and Ji, Heng and Jiang, Nan and Zhang, Tong},
  journal={arXiv preprint arXiv:2312.11456},
  year={2023}
}

@article{ethayarajh2110understanding,
  title={Understanding dataset difficulty with V-usable information (2021)},
  author={Ethayarajh, Kawin and Choi, Yejin and Swayamdipta, Swabha},
  journal={URL https://arxiv. org/abs/2110.08420}
}

@article{das2024active,
  title={Active preference optimization for sample efficient RLHF},
  author={Das, Nirjhar and Chakraborty, Souradip and Pacchiano, Aldo and Chowdhury, Sayak Ray},
  journal={arXiv preprint arXiv:2402.10500},
  year={2024}
}

@article{mehta2023sample,
  title={Sample efficient reinforcement learning from human feedback via active exploration},
  author={Mehta, Viraj and Das, Vikramjeet and Neopane, Ojash and Dai, Yijia and Bogunovic, Ilija and Schneider, Jeff and Neiswanger, Willie},
  journal={arXiv preprint arXiv:2312.00267},
  year={2023}
}

@article{yue2012k,
  title={The k-armed dueling bandits problem},
  author={Yue, Yisong and Broder, Josef and Kleinberg, Robert and Joachims, Thorsten},
  journal={Journal of Computer and System Sciences},
  volume={78},
  number={5},
  pages={1538--1556},
  year={2012},
  publisher={Elsevier}
}

@article{ziegler2019fine,
  title={Fine-tuning language models from human preferences},
  author={Ziegler, Daniel M and Stiennon, Nisan and Wu, Jeffrey and Brown, Tom B and Radford, Alec and Amodei, Dario and Christiano, Paul and Irving, Geoffrey},
  journal={arXiv preprint arXiv:1909.08593},
  year={2019}
}

@article{wang2024interpretable,
  title={Interpretable preferences via multi-objective reward modeling and mixture-of-experts},
  author={Wang, Haoxiang and Xiong, Wei and Xie, Tengyang and Zhao, Han and Zhang, Tong},
  journal={arXiv preprint arXiv:2406.12845},
  year={2024}
}

@article{llama3modelcard,
  title  = {Llama 3 Model Card},
  author = {AI@Meta},
  year   = {2024},
  url    = {https://github.com/meta-llama/llama3/blob/main/MODEL_CARD.md}
}

@article{Jiang2023Mistral7,
  title   = {Mistral {7B}},
  author  = {Albert Qiaochu Jiang and Alexandre Sablayrolles and Arthur Mensch and Chris Bamford and Devendra Singh Chaplot and Diego de Las Casas and Florian Bressand and Gianna Lengyel and Guillaume Lample and Lucile Saulnier and L'elio Renard Lavaud and Marie-Anne Lachaux and Pierre Stock and Teven Le Scao and Thibaut Lavril and Thomas Wang and Timoth{\'e}e Lacroix and William El Sayed},
  journal = {ArXiv},
  year    = {2023},
  volume  = {abs/2310.06825}
}

@article{team2024gemma,
  title   = {Gemma 2: Improving open language models at a practical size},
  author  = {Team, Gemma and Riviere, Morgane and Pathak, Shreya and Sessa, Pier Giuseppe and Hardin, Cassidy and Bhupatiraju, Surya and Hussenot, L{\'e}onard and Mesnard, Thomas and Shahriari, Bobak and Ram{\'e}, Alexandre and others},
  journal = {arXiv preprint arXiv:2408.00118},
  year    = {2024}
}

@article{wu2024alpha,
  title={Alpha-DPO: Adaptive Reward Margin is What Direct Preference Optimization Needs},
  author={Wu, Junkang and Wang, Xue and Yang, Zhengyi and Wu, Jiancan and Gao, Jinyang and Ding, Bolin and Wang, Xiang and He, Xiangnan},
  journal={arXiv preprint arXiv:2410.10148},
  year={2024}
}

@inproceedings{Ouyang2022TrainingLM,
  title     = {Training language models to follow instructions with human feedback},
  author    = {Long Ouyang and Jeff Wu and Xu Jiang and Diogo Almeida and Carroll L. Wainwright and Pamela Mishkin and Chong Zhang and Sandhini Agarwal and Katarina Slama and Alex Ray and John Schulman and Jacob Hilton and Fraser Kelton and Luke E. Miller and Maddie Simens and Amanda Askell and Peter Welinder and Paul Francis Christiano and Jan Leike and Ryan J. Lowe},
  booktitle = {NeurIPS},
  year      = {2022}
}

@article{zhou2024lima,
  title   = {{LIMA}: Less is more for alignment},
  author  = {Zhou, Chunting and Liu, Pengfei and Xu, Puxin and Iyer, Srinivasan and Sun, Jiao and Mao, Yuning and Ma, Xuezhe and Efrat, Avia and Yu, Ping and Yu, Lili and others},
  journal = {NeurIPS},
  year    = {2023}
}

@misc{taori2023stanford,
  title  = {Stanford alpaca: An instruction-following llama model},
  author = {Taori, Rohan and Gulrajani, Ishaan and Zhang, Tianyi and Dubois, Yann and Li, Xuechen and Guestrin, Carlos and Liang, Percy and Hashimoto, Tatsunori B},
  year   = {2023}
}

@article{geng2023koala,
  title   = {Koala: A dialogue model for academic research},
  author  = {Geng, Xinyang and Gudibande, Arnav and Liu, Hao and Wallace, Eric and Abbeel, Pieter and Levine, Sergey and Song, Dawn},
  journal = {Blog post, April},
  volume  = {1},
  pages   = {6},
  year    = {2023}
}

@online{DatabricksBlog2023DollyV2,
  author  = {Mike Conover and Matt Hayes and Ankit Mathur and Jianwei Xie and Jun Wan and Sam Shah and Ali Ghodsi and Patrick Wendell and Matei Zaharia and Reynold Xin},
  title   = {Free Dolly: Introducing the World's First Truly Open Instruction-Tuned {LLM}},
  year    = {2023},
  url     = {https://www.databricks.com/blog/2023/04/12/dolly-first-open-commercially-viable-instruction-tuned-llm},
  urldate = {2023-06-30}
}

@inproceedings{kopf2024openassistant,
  title     = {OpenAssistant Conversations-Democratizing Large Language Model Alignment},
  author    = {K{\"o}pf, Andreas and Kilcher, Yannic and von R{\"u}tte, Dimitri and Anagnostidis, Sotiris and Tam, Zhi Rui and Stevens, Keith and Barhoum, Abdullah and Nguyen, Duc Minh and Stanley, Oliver and Nagyfi, Rich{\'a}rd and others},
  booktitle = {Thirty-seventh Conference on Neural Information Processing Systems Datasets and Benchmarks Track},
  year      = {2023}
}

@inproceedings{Ding2023EnhancingCL,
  title     = {Enhancing Chat Language Models by Scaling High-quality Instructional Conversations},
  author    = {Ning Ding and Yulin Chen and Bokai Xu and Yujia Qin and Zhi Zheng and Shengding Hu and Zhiyuan Liu and Maosong Sun and Bowen Zhou},
  booktitle = {EMNLP},
  year      = {2023}
}

@inproceedings{wang2024openchat,
  title     = {{OpenChat}: Advancing Open-source Language Models with Mixed-Quality Data},
  author    = {Guan Wang and Sijie Cheng and Xianyuan Zhan and Xiangang Li and Sen Song and Yang Liu},
  booktitle = {ICLR},
  year      = {2024},
  url       = {https://openreview.net/forum?id=AOJyfhWYHf}
}

@inproceedings{chen2024alpagasus,
  title     = {{AlpaGasus}: Training a Better {Alpaca} with Fewer Data},
  author    = {Lichang Chen and Shiyang Li and Jun Yan and Hai Wang and Kalpa Gunaratna and Vikas Yadav and Zheng Tang and Vijay Srinivasan and Tianyi Zhou and Heng Huang and Hongxia Jin},
  booktitle = {ICLR},
  year      = {2024},
  url       = {https://openreview.net/forum?id=FdVXgSJhvz}
}

@inproceedings{gao2023scaling,
  title        = {Scaling laws for reward model overoptimization},
  author       = {Gao, Leo and Schulman, John and Hilton, Jacob},
  booktitle    = {International Conference on Machine Learning},
  pages        = {10835--10866},
  year         = {2023},
  organization = {PMLR}
}

@article{luo2023wizardmath,
  title   = {Wizardmath: Empowering mathematical reasoning for large language models via reinforced evol-instruct},
  author  = {Luo, Haipeng and Sun, Qingfeng and Xu, Can and Zhao, Pu and Lou, Jianguang and Tao, Chongyang and Geng, Xiubo and Lin, Qingwei and Chen, Shifeng and Zhang, Dongmei},
  journal = {arXiv preprint arXiv:2308.09583},
  year    = {2023}
}

@article{chen2024odin,
  title   = {{ODIN}: Disentangled Reward Mitigates Hacking in {RLHF}},
  author  = {Chen, Lichang and Zhu, Chen and Soselia, Davit and Chen, Jiuhai and Zhou, Tianyi and Goldstein, Tom and Huang, Heng and Shoeybi, Mohammad and Catanzaro, Bryan},
  journal = {arXiv preprint arXiv:2402.07319},
  year    = {2024}
}

@article{lightman2023let,
  title   = {Let's Verify Step by Step},
  author  = {Lightman, Hunter and Kosaraju, Vineet and Burda, Yura and Edwards, Harri and Baker, Bowen and Lee, Teddy and Leike, Jan and Schulman, John and Sutskever, Ilya and Cobbe, Karl},
  journal = {arXiv preprint arXiv:2305.20050},
  year    = {2023}
}

@article{havrilla2024glore,
  title   = {{GLoRe}: When, Where, and How to Improve {LLM} Reasoning via Global and Local Refinements},
  author  = {Havrilla, Alex and Raparthy, Sharath and Nalmpantis, Christoforus and Dwivedi-Yu, Jane and Zhuravinskyi, Maksym and Hambro, Eric and Railneau, Roberta},
  journal = {arXiv preprint arXiv:2402.10963},
  year    = {2024}
}

@article{lambert2024rewardbench,
  title   = {{RewardBench}: Evaluating Reward Models for Language Modeling},
  author  = {Nathan Lambert and Valentina Pyatkin and Jacob Daniel Morrison and Lester James Validad Miranda and Bill Yuchen Lin and Khyathi Raghavi Chandu and Nouha Dziri and Sachin Kumar and Tom Zick and Yejin Choi and Noah A. Smith and Hanna Hajishirzi},
  journal = {ArXiv},
  year    = {2024},
  volume  = {abs/2403.13787}
}

@article{amini2024direct,
  title   = {Direct Preference Optimization with an Offset},
  author  = {Amini, Afra and Vieira, Tim and Cotterell, Ryan},
  journal = {arXiv preprint arXiv:2402.10571},
  year    = {2024}
}

@inproceedings{korbak2023pretraining,
  title        = {Pretraining language models with human preferences},
  author       = {Korbak, Tomasz and Shi, Kejian and Chen, Angelica and Bhalerao, Rasika Vinayak and Buckley, Christopher and Phang, Jason and Bowman, Samuel R and Perez, Ethan},
  booktitle    = {International Conference on Machine Learning},
  pages        = {17506--17533},
  year         = {2023},
  organization = {PMLR}
}

@inproceedings{Zheng2023ClickCT,
  title     = {Click: Controllable Text Generation with Sequence Likelihood Contrastive Learning},
  author    = {Chujie Zheng and Pei Ke and Zheng Zhang and Minlie Huang},
  booktitle = {Findings of ACL},
  year      = {2023}
}

@article{dai2023safe,
  title   = {Safe {RLHF}: Safe reinforcement learning from human feedback},
  author  = {Dai, Josef and Pan, Xuehai and Sun, Ruiyang and Ji, Jiaming and Xu, Xinbo and Liu, Mickel and Wang, Yizhou and Yang, Yaodong},
  journal = {arXiv preprint arXiv:2310.12773},
  year    = {2023}
}

@inproceedings{tian2024finetuning,
  title     = {Fine-Tuning Language Models for Factuality},
  author    = {Katherine Tian and Eric Mitchell and Huaxiu Yao and Christopher D Manning and Chelsea Finn},
  booktitle = {The Twelfth International Conference on Learning Representations},
  year      = {2024},
  url       = {https://openreview.net/forum?id=WPZ2yPag4K}
}

@inproceedings{Wang2024ArithmeticCO,
  title   = {Arithmetic Control of {LLMs} for Diverse User Preferences: Directional Preference Alignment with Multi-Objective Rewards},
  author  = {Haoxiang Wang and Yong Lin and Wei Xiong and Rui Yang and Shizhe Diao and Shuang Qiu and Han Zhao and Tong Zhang},
  booktitle = {ACL},
  year    = {2024}
}

@article{nakano2021webgpt,
  title   = {{WebGPT}: Browser-assisted question-answering with human feedback},
  author  = {Nakano, Reiichiro and Hilton, Jacob and Balaji, Suchir and Wu, Jeff and Ouyang, Long and Kim, Christina and Hesse, Christopher and Jain, Shantanu and Kosaraju, Vineet and Saunders, William and others},
  journal = {arXiv preprint arXiv:2112.09332},
  year    = {2021}
}

@article{havrilla2024teaching,
  title   = {Teaching large language models to reason with reinforcement learning},
  author  = {Havrilla, Alex and Du, Yuqing and Raparthy, Sharath Chandra and Nalmpantis, Christoforos and Dwivedi-Yu, Jane and Zhuravinskyi, Maksym and Hambro, Eric and Sukhbaatar, Sainbayar and Raileanu, Roberta},
  journal = {arXiv preprint arXiv:2403.04642},
  year    = {2024}
}

@article{casper2023open,
  title   = {Open problems and fundamental limitations of reinforcement learning from human feedback},
  author  = {Casper, Stephen and Davies, Xander and Shi, Claudia and Gilbert, Thomas Krendl and Scheurer, J{\'e}r{\'e}my and Rando, Javier and Freedman, Rachel and Korbak, Tomasz and Lindner, David and Freire, Pedro and others},
  journal = {arXiv preprint arXiv:2307.15217},
  year    = {2023}
}

@article{singhal2023long,
  title   = {A Long Way to Go: Investigating Length Correlations in {RLHF}},
  author  = {Singhal, Prasann and Goyal, Tanya and Xu, Jiacheng and Durrett, Greg},
  journal = {arXiv preprint arXiv:2310.03716},
  year    = {2023}
}

@inproceedings{wang2023far,
  title     = {How Far Can Camels Go? Exploring the State of Instruction Tuning on Open Resources},
  author    = {Wang, Yizhong and Ivison, Hamish and Dasigi, Pradeep and Hessel, Jack and Khot, Tushar and Chandu, Khyathi and Wadden, David and MacMillan, Kelsey and Smith, Noah A and Beltagy, Iz and others},
  booktitle = {Thirty-seventh Conference on Neural Information Processing Systems Datasets and Benchmarks Track},
  year      = {2023}
}

@inproceedings{Rafailov2023DirectPO,
  title     = {Direct Preference Optimization: Your Language Model is Secretly a Reward Model},
  author    = {Rafael Rafailov and Archit Sharma and Eric Mitchell and Stefano Ermon and Christopher D. Manning and Chelsea Finn},
  booktitle = {NeurIPS},
  year      = {2023}
}

@article{Zhao2023SLiCHFSL,
  title   = {{SLiC-HF}: Sequence Likelihood Calibration with Human Feedback},
  author  = {Yao Zhao and Rishabh Joshi and Tianqi Liu and Misha Khalman and Mohammad Saleh and Peter J. Liu},
  journal = {ArXiv},
  year    = {2023},
  volume  = {abs/2305.10425}
}

@inproceedings{liu2024statistical,
  title     = {Statistical Rejection Sampling Improves Preference Optimization},
  author    = {Tianqi Liu and Yao Zhao and Rishabh Joshi and Misha Khalman and Mohammad Saleh and Peter J Liu and Jialu Liu},
  booktitle = {The Twelfth International Conference on Learning Representations},
  year      = {2024},
  url       = {https://openreview.net/forum?id=xbjSwwrQOe}
}

@article{dong2024rlhf,
  title   = {{RLHF} workflow: From reward modeling to online {RLHF}},
  author  = {Dong, Hanze and Xiong, Wei and Pang, Bo and Wang, Haoxiang and Zhao, Han and Zhou, Yingbo and Jiang, Nan and Sahoo, Doyen and Xiong, Caiming and Zhang, Tong},
  journal = {arXiv preprint arXiv:2405.07863},
  year    = {2024}
}

@article{Kim2024sDPODU,
  title   = {{sDPO}: Don't Use Your Data All at Once},
  author  = {Dahyun Kim and Yungi Kim and Wonho Song and Hyeonwoo Kim and Yunsu Kim and Sanghoon Kim and Chanjun Park},
  journal = {ArXiv},
  year    = {2024},
  volume  = {abs/2403.19270}
}

@article{Rosset2024DirectNO,
  title   = {Direct Nash Optimization: Teaching Language Models to Self-Improve with General Preferences},
  author  = {Corby Rosset and Ching-An Cheng and Arindam Mitra and Michael Santacroce and Ahmed Awadallah and Tengyang Xie},
  journal = {ArXiv},
  year    = {2024},
  volume  = {abs/2404.03715}
}

@inproceedings{xiong2024iterative,
  title     = {Iterative preference learning from human feedback: Bridging theory and practice for {RLHF} under {KL}-constraint},
  author    = {Xiong, Wei and Dong, Hanze and Ye, Chenlu and Wang, Ziqi and Zhong, Han and Ji, Heng and Jiang, Nan and Zhang, Tong},
  booktitle = {Forty-first International Conference on Machine Learning},
  year      = {2024}
}

@article{yuan2024self,
  title   = {Self-rewarding language models},
  author  = {Yuan, Weizhe and Pang, Richard Yuanzhe and Cho, Kyunghyun and Sukhbaatar, Sainbayar and Xu, Jing and Weston, Jason},
  journal = {arXiv preprint arXiv:2401.10020},
  year    = {2024}
}

@article{dong2023raft,
  title   = {{RAFT}: Reward rAnked FineTuning for Generative Foundation Model Alignment},
  author  = {Dong, Hanze and Xiong, Wei and Goyal, Deepanshu and Zhang, Yihan and Chow, Winnie and Pan, Rui and Diao, Shizhe and Zhang, Jipeng and KaShun, SHUM and Zhang, Tong},
  journal = {Transactions on Machine Learning Research},
  year    = {2023}
}

@article{liu2024lipo,
  title   = {{LiPO}: Listwise Preference Optimization through Learning-to-Rank},
  author  = {Liu, Tianqi and Qin, Zhen and Wu, Junru and Shen, Jiaming and Khalman, Misha and Joshi, Rishabh and Zhao, Yao and Saleh, Mohammad and Baumgartner, Simon and Liu, Jialu and others},
  journal = {arXiv preprint arXiv:2402.01878},
  year    = {2024}
}

@inproceedings{song2024preference,
  title     = {Preference ranking optimization for human alignment},
  author    = {Song, Feifan and Yu, Bowen and Li, Minghao and Yu, Haiyang and Huang, Fei and Li, Yongbin and Wang, Houfeng},
  booktitle = {AAAI},
  year      = {2024}
}

@article{Hong2024ORPOMP,
  title   = {{ORPO}: Monolithic Preference Optimization without Reference Model},
  author  = {Jiwoo Hong and Noah Lee and James Thorne},
  journal = {ArXiv},
  year    = {2024},
  volume  = {abs/2403.07691}
}

@article{xu2023some,
  title   = {Some things are more cringe than others: Preference optimization with the pairwise cringe loss},
  author  = {Xu, Jing and Lee, Andrew and Sukhbaatar, Sainbayar and Weston, Jason},
  journal = {arXiv preprint arXiv:2312.16682},
  year    = {2023}
}

@article{bansal2024comparing,
  title   = {Comparing Bad Apples to Good Oranges: Aligning Large Language Models via Joint Preference Optimization},
  author  = {Bansal, Hritik and Suvarna, Ashima and Bhatt, Gantavya and Peng, Nanyun and Chang, Kai-Wei and Grover, Aditya},
  journal = {arXiv preprint arXiv:2404.00530},
  year    = {2024}
}

@article{zheng2024weak,
  title   = {Weak-to-strong extrapolation expedites alignment},
  author  = {Zheng, Chujie and Wang, Ziqi and Ji, Heng and Huang, Minlie and Peng, Nanyun},
  journal = {arXiv preprint arXiv:2404.16792},
  year    = {2024}
}

@article{Azar2023AGT,
  title   = {A General Theoretical Paradigm to Understand Learning from Human Preferences},
  author  = {Mohammad Gheshlaghi Azar and Mark Rowland and Bilal Piot and Daniel Guo and Daniele Calandriello and Michal Valko and R{\'e}mi Munos},
  journal = {ArXiv},
  year    = {2023},
  volume  = {abs/2310.12036}
}

@article{Ethayarajh2024KTOMA,
  title   = {{KTO}: Model Alignment as Prospect Theoretic Optimization},
  author  = {Kawin Ethayarajh and Winnie Xu and Niklas Muennighoff and Dan Jurafsky and Douwe Kiela},
  journal = {ArXiv},
  year    = {2024},
  volume  = {abs/2402.01306}
}

@article{Park2024DisentanglingLF,
  title   = {Disentangling Length from Quality in Direct Preference Optimization},
  author  = {Ryan Park and Rafael Rafailov and Stefano Ermon and Chelsea Finn},
  journal = {ArXiv},
  year    = {2024},
  volume  = {abs/2403.19159}
}

@article{anthony2017thinking,
  title   = {Thinking fast and slow with deep learning and tree search},
  author  = {Anthony, Thomas and Tian, Zheng and Barber, David},
  journal = {Advances in neural information processing systems},
  volume  = {30},
  year    = {2017}
}

@article{gao2024impact,
  title={Impact of preference noise on the alignment performance of generative language models},
  author={Gao, Yang and Alon, Dana and Metzler, Donald},
  journal={arXiv preprint arXiv:2404.09824},
  year={2024}
}

@inproceedings{ban2021local,
  title={Local clustering in contextual multi-armed bandits},
  author={Ban, Yikun and He, Jingrui},
  booktitle={Proceedings of the Web Conference 2021},
  pages={2335--2346},
  year={2021}
}

@inproceedings{qi2023graph,
  title={Graph neural bandits},
  author={Qi, Yunzhe and Ban, Yikun and He, Jingrui},
  booktitle={Proceedings of the 29th ACM SIGKDD Conference on Knowledge Discovery and Data Mining},
  pages={1920--1931},
  year={2023}
}

@inproceedings{ban2024meta,
  title={Meta clustering of neural bandits},
  author={Ban, Yikun and Qi, Yunzhe and Wei, Tianxin and Liu, Lihui and He, Jingrui},
  booktitle={Proceedings of the 30th ACM SIGKDD Conference on Knowledge Discovery and Data Mining},
  pages={95--106},
  year={2024}
}

@article{he2024llm,
  title={LLM-Forest: Ensemble Learning of LLMs with Graph-Augmented Prompts for Data Imputation},
  author={He, Xinrui and Ban, Yikun and Zou, Jiaru and Wei, Tianxin and Cook, Curtiss B and He, Jingrui},
  journal={arXiv preprint arXiv:2410.21520},
  year={2024}
}

@article{zou2025transformer,
  title={Transformer Copilot: Learning from The Mistake Log in LLM Fine-tuning},
  author={Zou, Jiaru and Ban, Yikun and Li, Zihao and Qi, Yunzhe and Qiu, Ruizhong and Yang, Ling and He, Jingrui},
  journal={arXiv preprint arXiv:2505.16270},
  year={2025}
}

@article{ban2022improved,
  title={Improved algorithms for neural active learning},
  author={Ban, Yikun and Zhang, Yuheng and Tong, Hanghang and Banerjee, Arindam and He, Jingrui},
  journal={Advances in Neural Information Processing Systems},
  volume={35},
  pages={27497--27509},
  year={2022}
}

@article{li2025interpretable,
  title={Interpretable unsupervised joint denoising and enhancement for real-world low-light scenarios},
  author={Li, Huaqiu and Hu, Xiaowan and Wang, Haoqian},
  journal={arXiv preprint arXiv:2503.14535},
  year={2025}
}

@inproceedings{li2025prompt,
  title={Prompt-sid: Learning structural representation prompt via latent diffusion for single image denoising},
  author={Li, Huaqiu and Zhang, Wang and Hu, Xiaowan and Jiang, Tao and Chen, Zikang and Wang, Haoqian},
  booktitle={Proceedings of the AAAI Conference on Artificial Intelligence},
  volume={39},
  number={5},
  pages={4734--4742},
  year={2025}
}

@inproceedings{li2025ld,
  title={Ld-rps: Zero-shot unified image restoration via latent diffusion recurrent posterior sampling},
  author={Li, Huaqiu and Wang, Yong and Huang, Tongwen and Huang, Hailang and Wang, Haoqian and Chu, Xiangxiang},
  booktitle={Proceedings of the IEEE/CVF International Conference on Computer Vision},
  pages={13684--13694},
  year={2025}
}

@article{fu2022sparsett,
  title={SparseTT: Visual tracking with sparse transformers},
  author={Fu, Zhihong and Fu, Zehua and Liu, Qingjie and Cai, Wenrui and Wang, Yunhong},
  journal={arXiv preprint arXiv:2205.03776},
  year={2022}
}

@inproceedings{cai2024hiptrack,
  title={Hiptrack: Visual tracking with historical prompts},
  author={Cai, Wenrui and Liu, Qingjie and Wang, Yunhong},
  booktitle={Proceedings of the IEEE/CVF Conference on Computer Vision and Pattern Recognition},
  pages={19258--19267},
  year={2024}
}

@article{cai2023learning,
  title={Learning historical status prompt for accurate and robust visual tracking},
  author={Cai, Wenrui and Liu, Qingjie and Wang, Yunhong},
  journal={arXiv preprint arXiv:2311.02072},
  volume={7},
  year={2023}
}

@inproceedings{cai2025spmtrack,
  title={SPMTrack: spatio-temporal parameter-efficient fine-tuning with mixture of experts for scalable visual tracking},
  author={Cai, Wenrui and Liu, Qingjie and Wang, Yunhong},
  booktitle={Proceedings of the computer vision and pattern recognition conference},
  pages={16871--16881},
  year={2025}
}

@article{cai2025seednorm,
  title={SeeDNorm: Self-Rescaled Dynamic Normalization},
  author={Cai, Wenrui and Zhu, Defa and Liu, Qingjie and Min, Qiyang},
  journal={arXiv preprint arXiv:2510.22777},
  year={2025}
}

@inproceedings{zhao2025redone,
  title={RedOne: Revealing Domain-specific LLM Post-Training in Social Networking Services},
  author={Zhao, Fei and Lu, Chonggang and Xie, Zheyong and Liu, Ziyan and Qian, Haofu and Huang, Jianzhao and Shi, Fangcheng and Meng, Zijie and Guo, Hongcheng and He, Mingqian and others},
  booktitle={Proceedings of the 2025 Conference on Empirical Methods in Natural Language Processing: Industry Track},
  pages={2648--2674},
  year={2025}
}

@inproceedings{guo2025pet,
  title={Pet-Bench: Benchmarking the Abilities of Large Language Models as E-Pets in Social Network Services},
  author={Guo, Hongcheng and Xie, Zheyong and Cao, Shaosheng and Wang, Boyang and Liu, Weiting and Ye, Zheyu and Li, Zhoujun and Liu, Zuozhu and Lu, Wei},
  booktitle={Proceedings of the 34th ACM International Conference on Information and Knowledge Management},
  pages={6402--6407},
  year={2025}
}

@article{huang2026does,
  title={Does Your Reasoning Model Implicitly Know When to Stop Thinking?},
  author={Huang, Zixuan and Xia, Xin and Ren, Yuxi and Zheng, Jianbin and Wang, Xuanda and Zhang, Zhixia and Xie, Hongyan and Liang, Songshi and Chen, Zehao and Xiao, Xuefeng and others},
  journal={arXiv preprint arXiv:2602.08354},
  year={2026}
}

@article{zhang2026heterogeneous,
  title={Heterogeneous Agent Collaborative Reinforcement Learning},
  author={Zhang, Zhixia and Huang, Zixuan and Xia, Xin and Wang, Deqing and Zhuang, Fuzhen and Ma, Shuai and Ding, Ning and Yang, Yaodong and Li, Jianxin and Ban, Yikun},
  journal={arXiv preprint arXiv:2603.02604},
  year={2026}
}

@article{huang2026real,
  title={Real-Time Aligned Reward Model beyond Semantics},
  author={Huang, Zixuan and Xia, Xin and Ren, Yuxi and Zheng, Jianbin and Xiao, Xuefeng and Xie, Hongyan and Huaqiu, Li and Liang, Songshi and Dai, Zhongxiang and Zhuang, Fuzhen and others},
  journal={arXiv preprint arXiv:2601.22664},
  year={2026}
}

\newpage
\begin{enumerate}

\item {\bf Claims}
    \item[] Question: Do the main claims made in the abstract and introduction accurately reflect the paper's contributions and scope?
    \item[] Answer: \answerYes{} 
    \item[] Justification: In the Section~\ref{intro}, we discuss the problems we identified, propose our methodology, and summarize our contributions.

\item {\bf Limitations}
    \item[] Question: Does the paper discuss the limitations of the work performed by the authors?
    \item[] Answer:  \answerYes{}
    \item[] Justification: We specifically discuss the limitations of this work in Appendix~\ref{limit}.

\item {\bf Theory assumptions and proofs}
    \item[] Question: For each theoretical result, does the paper provide the full set of assumptions and a complete (and correct) proof?
    \item[] Answer: \answerNA{} 
    \item[] Justification: Our work does not involve specific theoretical analysis.

    \item {\bf Experimental result reproducibility}
    \item[] Question: Does the paper fully disclose all the information needed to reproduce the main experimental results of the paper to the extent that it affects the main claims and/or conclusions of the paper (regardless of whether the code and data are provided or not)?
    \item[] Answer: \answerYes{} 
    \item[] Justification: We provide complete experimental details and parameter settings in the  the Appendix~\ref{detail}.

\item {\bf Open access to data and code}
    \item[] Question: Does the paper provide open access to the data and code, with sufficient instructions to faithfully reproduce the main experimental results, as described in supplemental material?
    \item[] Answer: \answerYes{} 
    \item[] Justification: We provide the complete code and the necessary instructions for running it in the supplementary materials.

\item {\bf Experimental setting/details}
    \item[] Question: Does the paper specify all the training and test details (e.g., data splits, hyperparameters, how they were chosen, type of optimizer, etc.) necessary to understand the results?
    \item[] Answer: \answerYes{} 
    \item[] Justification:  We provide complete training and test details and parameter settings in the Appendix~\ref{detail}.

\item {\bf Experiment statistical significance}
    \item[] Question: Does the paper report error bars suitably and correctly defined or other appropriate information about the statistical significance of the experiments?
    \item[] Answer: \answerYes{} 
    \item[] Justification: All reported results are obtained by taking the average of runs with 10 random seeds.

\item {\bf Experiments compute resources}
    \item[] Question: For each experiment, does the paper provide sufficient information on the computer resources (type of compute workers, memory, time of execution) needed to reproduce the experiments?
    \item[] Answer: \answerYes{} 
    \item[] Justification: We analyze the running time and computational cost of our method in the Section~\ref{compute}, and provide the computational resources we use, which is in the Appendix~\ref{device}.
    
\item {\bf Code of ethics}
    \item[] Question: Does the research conducted in the paper conform, in every respect, with the NeurIPS Code of Ethics \url{https://neurips.cc/public/EthicsGuidelines}?
    \item[] Answer: \answerYes{}{} 
    \item[] Justification: We have reviewed the NeurIPS Code of Ethics and strictly adhere to any provisions therein.

\item {\bf Broader impacts}
    \item[] Question: Does the paper discuss both potential positive societal impacts and negative societal impacts of the work performed?
    \item[] Answer: \answerYes{} 
    \item[] Justification: There is no societal impact of our work performed.
    
\item {\bf Safeguards}
    \item[] Question: Does the paper describe safeguards that have been put in place for responsible release of data or models that have a high risk for misuse (e.g., pretrained language models, image generators, or scraped datasets)?
    \item[] Answer:\answerNA{} 
    \item[] Justification: Our work poses no such risks.

\item {\bf Licenses for existing assets}
    \item[] Question: Are the creators or original owners of assets (e.g., code, data, models), used in the paper, properly credited and are the license and terms of use explicitly mentioned and properly respected?
    \item[] Answer: \answerYes{}
    \item[] Justification: We comply with all the requirements mentioned in the guidelines.

\item {\bf New assets}
    \item[] Question: Are new assets introduced in the paper well documented and is the documentation provided alongside the assets?
    \item[] Answer: \answerNA{} 
    \item[] Justification: Our work does not release new assets.

\item {\bf Crowdsourcing and research with human subjects}
    \item[] Question: For crowdsourcing experiments and research with human subjects, does the paper include the full text of instructions given to participants and screenshots, if applicable, as well as details about compensation (if any)? 
    \item[] Answer: \answerNA{} 
    \item[] Justification: Our work does not involve crowdsourcing nor research with human subjects.

\item {\bf Institutional review board (IRB) approvals or equivalent for research with human subjects}
    \item[] Question: Does the paper describe potential risks incurred by study participants, whether such risks were disclosed to the subjects, and whether Institutional Review Board (IRB) approvals (or an equivalent approval/review based on the requirements of your country or institution) were obtained?
    \item[] Answer: \answerNA{}
    \item[] Justification: Our work does not involve crowdsourcing nor research with human subjects.

\item {\bf Declaration of LLM usage}
    \item[] Question: Does the paper describe the usage of LLMs if it is an important, original, or non-standard component of the core methods in this research? Note that if the LLM is used only for writing, editing, or formatting purposes and does not impact the core methodology, scientific rigorousness, or originality of the research, declaration is not required.
    \item[] Answer: \answerYes{} 
    \item[] Justification: We use LLMs as the judge model to audit the preference dataset, with specific details provided in the Appendix~\ref{supp:gptjudge}.

\end{enumerate}

\newpage

\appendix

\textbf{\Large Appendix}

\renewcommand{\contentsname}{Table of Contents}

\addtocontents{toc}{\protect\setcounter{tocdepth}{2}}

\tableofcontents

\newpage

\section{Limitations} \label{limit}
The primary limitation of SamS lies in its performance sensitivity to data quality. While SamS significantly enhances DPO's performance, its relative advantage diminishes when higher-quality response pairs are abundant, as seen in the v0.2 setting. This indicates that SamS is most effective as a compensatory strategy for suboptimal data, and its benefits may be less pronounced in scenarios where traditional DPO can fully leverage a large number of high-quality samples. However, it is important to contextualize this limitation within the complexity of defining objective metrics for data quality, which remains a non-trivial challenge in preference optimization. Moreover, this constraint may be mitigated by integrating SamS with data pre-selection strategies, as demonstrated in Appendix~\ref{selective dpo}.

\section{Broader Impact}
Our proposed SamS offers several significant advantages and has far-reaching potential applications. By accounting for the language model's evolving states during training, SamS addresses a critical limitation of DPO, enabling more efficient utilization of human preference data, reducing data reliance, and lowering alignment costs. Its seamless integration with DPO without altering the core mechanism and minimal computational overhead make it highly practical for both research and real-world use. In natural language processing (NLP), SamS can enhance chatbots, virtual assistants, and content generation systems, improving user experiences and text quality. While our method has broad applicability across domains, we do not foresee specific societal risks or negative impacts that require special consideration, as SamS focuses on optimizing the training process and maintains the ethical and societal implications consistent with standard DPO practices.

\section{Additional Related Work}

\paragraph{Reinforcement learning from human feedback.} 
RLHF is a critical technique for aligning large language models with human preferences~\cite{christiano2017deep, ziegler2019fine, Ouyang2022TrainingLM, bai2022training}. The classical RLHF pipeline typically comprises three phases: supervised fine-tuning~\cite{zhou2024lima,taori2023stanford, geng2023koala, DatabricksBlog2023DollyV2, kopf2024openassistant, Ding2023EnhancingCL, wang2024openchat, chen2024alpagasus, xia2024less}, reward model training~\cite{gao2023scaling,luo2023wizardmath, chen2024odin,lightman2023let,havrilla2024glore, lambert2024rewardbench}, and policy optimization against the reward model~\cite{schulman2017proximal, anthony2017thinking}. As a classic reinforcement learning algorithm,  Proximal Policy Optimization (PPO)~\cite{schulman2017proximal} is widely used in the third stage of RLHF. The RLHF framework is extensively utilized across a range of applications, such as mitigating toxicity~\cite{amini2024direct,korbak2023pretraining,Zheng2023ClickCT}, ensuring safety~\cite{dai2023safe}, enhancing helpfulness~\cite{tian2024finetuning,Wang2024ArithmeticCO}, searching and navigating the web~\cite{nakano2021webgpt}, and improving model reasoning abilities~\cite{havrilla2024teaching}. Recently,~\cite{casper2023open} has identified challenges throughout the RLHF pipeline, spanning preference data collection to model training. Additional studies have shown that RLHF may result in biased outcomes, including overly verbose model outputs~\cite{dubois2024length, singhal2023long, wang2023far}.
Subsequent algorithm variants have continued to emerge and provided multifaceted optimizations for them \citep{zhang2026heterogeneous, huang2026does, huang2026real}.
In addition to the methods discussed above, a wide range of advanced techniques have been proposed in recent years to address various challenges in representation learning, model optimization, and generative modeling. These include progress in interpretable representation learning~\cite{li2025interpretable}, prompt-based structural modeling~\cite{li2025prompt}, diffusion-driven restoration~\cite{li2025ld}, efficient transformer architectures for visual modeling~\cite{fu2022sparsett}, prompt-guided sequence modeling~\cite{cai2023learning,cai2024hiptrack}, parameter-efficient tuning strategies~\cite{cai2025spmtrack}, as well as novel normalization mechanisms for improving model stability~\cite{cai2025seednorm}.
Although these works are designed for different task scenarios, they collectively enrich the toolkit of modern machine learning research and provide useful insights for understanding the generalization and optimization of neural models.

\paragraph{Difference from Existing Related Problems.}
Several related problem settings exist, which we outline and analyze here to highlight their differences from our Sample Scheduling problem:

\textbf{(1) Active Human Feedback Collection for DPO}.
   Based on Online Iterative DPO~\cite{xiong2023iterative}, this setting includes studies such as~\cite{das2024active, muldrew2024active, ji2024reinforcement}. These methods actively select prompts $x_{t,i}$ from a dataset, generate responses online during training, and subsequently have these responses annotated by an oracle to form pairs $(y^w_{t,i}, y^l_{t,i})$. Unlike our method, their primary goal is to optimize query quality given a fixed annotation budget.
   
\textbf{(2) Contextual Dueling Bandits for DPO}.
Studies that include~\cite{mehta2023sample, liu2024sample} adopt the online iterative DPO framework, describing the selection of the response pair as a contextual dueling bandit problem~\cite{yue2012k, dudik2015contextual}. These approaches use exploration-exploitation to select response pairs for preference datasets, while our method applies such principles to sample scheduling in each training round.

\textbf{(3) Data Selection for DPO}.
   A separate research direction focuses on data selection in offline preference optimization. For instance,~\cite{shen2024towards} conducts a fine-grained analysis of preference data and proposes evaluation metrics. Similarly, ~\cite{deng2025less,shen2024towards, deng2025less, gao2025principled} presents sample-quality evaluation approaches based on different observations, subsequently selecting data subsets for policy training. Although these methodologies train policies on selected subsets, they isolate the sample selection from the model's training process, thereby disregarding the dynamic interaction between selected samples and the evolving state of the model. This category essentially focuses on data preprocessing.

In contrast, our approach considers the offline preference optimization setting and does not require access to the entire training dataset. The scheduler in our framework dynamically and interactively selects samples during the training process of the policy $\pi_{\theta}$, guided explicitly by the evolving internal states of $\pi_{\theta}$. This dynamic sample scheduling establishes a novel reinforcement learning paradigm.

\section{Experimental Details} \label{detail}

In this section, we first provide a detailed description of the experimental setup, including the hyperparameters of the scheduler and the training and evaluation settings employed. Next, we compare SamS with Data Pre-Selection methods, which are the most related to our problem setting. Finally, we conduct an ablation study on the scheduler selection ratio and the Exploration Network $f^{S'}$.

\pagestyle{empty}
\begin{table}[ht]
\centering
\small
\caption{Evaluation details for AlpacaEval 2~\cite{dubois2024length} and MT-Bench~\cite{zheng2023judging}. Exs denotes the number of test examples. For AlpacaEval 2, LC refers to the length-controlled win rate~\cite{dubois2024length}, which mitigates the bias of judge models favoring longer responses.} 
\vspace{10pt}

\label{tab:benchmark} 
\setlength{\tabcolsep}{3pt} 
\begin{tabular}{@{}lccccc@{}}
\toprule
 & \textbf{\# Exs}. & \textbf{Baseline Model} & \textbf{Judge Model} & \textbf{Scoring Type} & \textbf{Metric} \\
\midrule
\textbf{AlpacaEval 2} & 805 & GPT-4 Turbo & GPT-4 Turbo & Pairwise comparison & LC \& raw win rate \\
\textbf{MT-Bench} & 80 & - & GPT-4 Turbo & Single-answer grading & Rating of 1 - 10 \\
\bottomrule
\end{tabular}
\end{table}

\subsection{Experimental Setup} \label{app:expset}
\paragraph{Scheduler Settings.}

For the encoder layer of $f$, we initialize it with \href{https://huggingface.co/sentence-transformers/all-MiniLM-L6-v2}{all-MiniLM-L6-v2}.

To improve the training efficiency, We pretrain the encoder layer offline and freeze its weights during the preference optimization process. The specific training details are provided in the Appendix~\ref{supp:pretrain}.
For the Exploitation Network $f^{S}$, we set its width $m = 4096$ and depth $L = 16$. As described in Section~\ref{sec:architecture}, we first concatenate the hidden states of $f^{S}$. Then, we perform downsampling using a parameter of 4, which entails calculating the average of every four consecutive positions. 
For the Exploration Network $f^{S'}$, we also set its depth $L=16$. Its width is jointly determined by the depth of $f^{S}$ and the downsampling parameter.
For Scheduler Training, We sample 32 offline batches from the random sample pool $\mathcal{P}$ at each round $t$, which has a capacity of 40,000.
We use the Adam optimizer for both $f^{S}$ and $f^{S'}$, and set the initial learning rate to $10^{-4}$. 
For Schedule Selection, we set the scheduling budget $|\widetilde{X}_t|=\frac{1}{2}|{X}_t|$.

\paragraph{Baselines.}

Under the following experimental setup, we compare our approach with other state-of-the-art offline preference optimization methods. Among these, RRHF~\cite{yuan2023rrhf} and SLiC-HF~\cite{zhao2023slic} both utilize ranking losses. RRHF employs a length-normalized log-likelihood function, whereas SLiC-HF~\cite{zhao2023slic} directly uses the log-likelihood function and incorporates an SFT objective. IPO~\cite{azar2024general} is a theoretically grounded method that avoids DPO's assumption that pairwise preferences can be substituted with pointwise rewards. CPO~\cite{xu2024contrastive} uses sequence likelihood as a reward and trains along the SFT objective. KTO~\cite{ethayarajh2024kto} learns from non-paired preference data. ORPO~\cite{hong2024orpo} introduces a reference-model-free odd ratio term to directly contrast winning and losing responses with the policy model and jointly trains with the SFT objective. R-DPO~\cite{park2024disentangling} is an enhanced version of DPO that incorporates an additional regularization term to mitigate length exploitation.

\paragraph{Preference Dataset Generation.}
To ensure fairness in comparisons, We adopt experimental settings that are currently widely used~\cite{meng2024simpo,wu2024alpha,hong2024orpo}. We utilize widely adopted instruction-tuned models as SFT models and employ the SFT model to generate five responses for each prompt $x$ in the UltraFeedback dataset~\cite{cui2023ultrafeedback}. Subsequently, a pretrained reward model serves as the annotator to directly assign a reward score $r(x,y_i)$ to each candidate response $y_i$. We then select the two responses with the largest score difference $y^w=y_{argmax(r)} $, $y^l=y_{argmin(r)}$ to form a sample $(x, y^w, y^l)$ in the preference dataset $\mathcal{D}$.

\paragraph{LLM Settings.}
We conduct experiments using two model settings.
The first model setting employs mistralai/Mistral-7B-Instruct-v0.2~\cite{Jiang2023Mistral7} and meta-llama/Meta-Llama-3-8B-Instruct~\cite{llama3modelcard} as SFT models, with llm-blender/PairRM~\cite{jiang2023llm} serving as the reward model.
The second model setting, which we refer to v0.2, employs meta-llama/Meta-Llama-3-8B-Instruct~\cite{llama3modelcard} and google/gemma-2-9b-it~\cite{team2024gemma} as SFT models. We utilize the more powerful RLHFlow/ArmoRM-Llama3-8B-v0.1~\cite{wang2024interpretable} as the reward model.
Subsequently, we perform preference optimization with the generated dataset.

\paragraph{Hyperparameters.}
We set the sampling temperature to 0.8 when generating responses with the SFT model. For DPO, we set $\beta = 0.01$, with a learning rate of $5 \times 10^{-7}$ for Mistral-7B-Instruct-v0.2, $1 \times 10^{-6}$ for Meta-Llama-3-8B-Instruct, and $3 \times 10^{-7}$ for gemma2-9b-it.

\paragraph{Evaluation Settings.} 
We primarily evaluate our models using two widely adopted open-ended instruction-following benchmarks: MT-Bench~\cite{zheng2023judging} and AlpacaEval 2~\cite{dubois2024length}. These benchmarks assess the models' general conversational capabilities across diverse query sets, with specific configurations detailed in Table~\ref{tab:benchmark}.
All the training experiments in this paper were conducted on 8 A100 GPUs.

\subsection{Dataset Details} \label{app:benchmarks}

Detailed information about the datasets used in the experiments is presented in Table~\ref{tab:data_detail}. For HH and SHP, we directly utilize the open-source data available on HuggingFace. For UltraFeedback, to ensure that the chosen responses in the training samples during preference optimization are in-distribution, we use only the prompts from the dataset and generate the offline preference dataset following the approach described in Appendix~\ref{app:expset}.

\begin{table}[ht]
    \centering
    \caption{Statistical information about the training datasets used in the experiments.}
    \vspace{10pt} 
    \resizebox{0.8\textwidth}{!}{
    \begin{tabular}{lccc}
        \toprule
        \textbf{Dataset} & $\boldsymbol{|\mathcal{D}_{train}|}$ & $\boldsymbol{|\mathcal{D}_{test}|}$ & \textbf{Type}\\
        \midrule
        HH & 160800 & 8552 & Helpful \& Harmless \\
        SHP & 348718 & 18409 & Hybrid \\
        UltraFeedback-Mistral & 56904 & 1866 & Hybrid \\
        UltraFeedback-Llama3 & 58119 & 1906 & Hybrid \\
        UltraFeedback-Llama3-v0.2 & 59876 & 1961 & Hybrid \\
        UltraFeedback-Gemma-v0.2 & 59569 & 1941 & Hybrid \\
        \bottomrule
    \end{tabular}
    }
    \label{tab:data_detail}
\end{table}

\subsection{Evaluation Details}

We provide a detailed version of Table~\ref{tab:exp1}, which is Table~\ref{tab:detail_exp1}.

\begin{table}[ht]
\centering
\caption{The evaluation metrics at the position where the policy converges.}
\vspace{10pt} 
\resizebox{0.8\textwidth}{!}{
\begin{tabular}{c c c c c}
\toprule
\textbf{Dataset} & \textbf{Method} & \textbf{Test-Acc($\%$)} & \textbf{Chosen Reward} & \textbf{Chosen Logps}\\
\midrule 
\multirow{6}{*}{\textbf{HH}} 
& \textbf{DPO} & 64.3 & -8.54 & -205 \\
& \textbf{DPO+SamS} & 67.1 & -5.52 & -176 \\
& \textbf{Improvement} & \textbf{+2.8} & \textbf{+35.36\%} & \textbf{+14.15\%} \\
\cmidrule{2-5}
& \textbf{KTO} & 60.2 & -0.404 & -287 \\
& \textbf{KTO+SamS} & 63.3 & -0.358 & -285 \\
& \textbf{Improvement} & \textbf{+3.1} & \textbf{+11.39\%} & \textbf{+0.7\%} \\
\midrule
\multirow{6}{*}{\textbf{SHP}} 
& \textbf{DPO} & 67.6 & -7.11 & -361 \\
& \textbf{DPO+SamS} & 70.0 & -5.64 & -341 \\
& \textbf{Improvement} & \textbf{+2.4} & \textbf{+20.68\%} & \textbf{+5.54\%} \\
\cmidrule{2-5}
& \textbf{KTO} & 65.2 & -1.22 & -134 \\
& \textbf{KTO+SamS} & 67.5 & -1.07 & -130 \\
& \textbf{Improvement} & \textbf{+2.3} & \textbf{+12.3\%} & \textbf{+2.99\%} \\
\bottomrule
\end{tabular}
}
\label{tab:detail_exp1}
\end{table}



\subsection{Ablation Study} \label{app:abstudy}

In this section, we conduct in-depth ablation studies to evaluate the effectiveness of the  scheduler selection ratio and the Exploration Network $f^{S'}$. 
Building upon the experimental setup described in Section~\ref{exp:epoch}, we utilize the Anthropic-HH dataset as the preference dataset and Pythia-2.8B as the foundation model, integrating SamS into  DPO.

\begin{figure}[ht]
  \centering
  \includegraphics[width=0.6\textwidth]{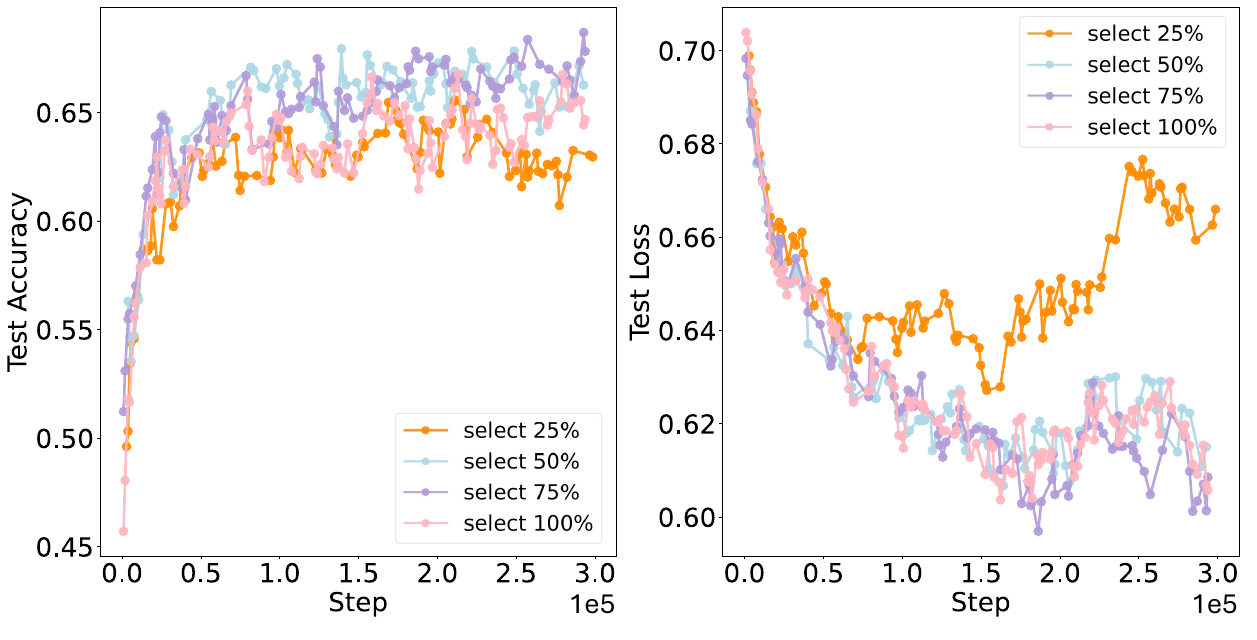}
  \caption{A comparison of different scheduler selection ratios in SamS reveals that 75\% outperforms 50\%, which in turn surpasses 100\%, followed by 25\%.}
  \label{fig:abstudy}
\end{figure}

To investigate the impact of different sample scheduling ratios, we let the scheduler select 25\%, 50\%, 75\%, and 100\% of the samples in each batch for the policy to learn (where selecting 100\% corresponds to standard DPO), as shown in the Figure~\ref{fig:abstudy}.

The results demonstrate that SamS significantly outperforms the original preference optimization method at higher sample selection ratios. Specifically, at scheduling ratios of 50\% and 75\%, SamS consistently achieves higher test accuracy than DPO. However, when SamS selects only 25\% of the samples, its performance noticeably declines compared to DPO, indicating that, with limited sample capacity, the potential gains from the small subset of samples scheduled by SamS for the policy are inferior to those from the entire batch.

To evaluate the function of the exploration network $f^{S'}$ within the scheduler, we conduct a comparative experiment with and without $f^{S'}$, as depicted in the  Table~\ref{tab:abstudy}. The results demonstrate that when $f^{S'}$ is included, SamS enables the policy to consistently achieve higher test accuracy, which demonstrates the necessity of $f^{S'}$.

\begin{table}[ht] 
 \centering
    \caption{Ablation study of SamS.}
    \vspace{10pt} 
    \begin{tabular}{lcc}
        \toprule
        \textbf{Method} & \textbf{Test-Acc}  \\
        \midrule
        Select 25\% & 63\%  \\
        Select 50\% & 67\%  \\
        Select 75\% & 68\% \\
        Select 100\% & 64\%  \\
        \midrule
        with $f^{S'}$ & 67\% \\
        without $f^{S'}$ & 65\%  \\
        \bottomrule
    \end{tabular}
    \label{tab:abstudy}
\end{table}

\section{GPT Judgement} \label{supp:gptjudge}
We employ GPT-4 Turbo as the judge model to rigorously evaluate a randomly sampled mixed subset of existing offline preference datasets. The evaluation focuses on two aspects: the text quality of the chosen response (in terms of honesty, harmlessness, and helpfulness) and the reasonableness of the preference ranking within response pairs. We randomly sample 2,000 instances from the Anthropic-HH dataset, Table~\ref{tab:judgedata} presents the results of this review. The findings indicate that, although preference rankings in existing offline datasets are generally accurate, the higher-preferred responses often exhibit issues such as dishonesty or potential harmfulness.

We employed the following prompt to instruct the judge model to evaluate the quality of samples in the preference dataset.
{\small \begin{spverbatim}
"""You are a human preference dataset evaluation expert. 
Now there is a manually labeled preference dataset sample. 
Each data entry consists of three fields: "prompt", "chosen_response", and "rejected_response", 
indicating that when the model is given the prompt, the chosen_response is more aligned with human preferences than the rejected_response.
Please evaluate the rationality of the preference dataset from the following aspects:
First, the text quality of the chosen_response, whether it fully meets the three requirements of being honest, harmless, and helpful.
Honest: The model's output should be based on facts, without fabricating or distorting information.
Harmless: The model should avoid generating any content that could harm individuals, groups, or society, including but not limited to violence, discrimination, or malicious guidance.
Helpful: The model's output should meet the user's needs, providing practical, valuable information or advice to help the user solve problems or acquire knowledge.

Second, determine whether there is an issue with the relative quality of the chosen_response and the rejected_response, 
i.e.,Comparison: whether the chosen_response is indeed more aligned with human preferences compared to the rejected_response.(4)

Based on the above two aspects, provide an evaluation. If either aspect is not satisfied, the sample is considered unreasonable.
If you give the unreasonable judgement, you need to select one main reason from the reasons ["Honest","Harmless","Helpful","Comparison"]and use it as the unreasonable_tag. 
If you give the reasonable judgemeny, you set the reason_tag to "null"
The content of the prompt is reasonable, and there is no need to consider the quality of the prompt. 

The returned text should be a JSON-formatted string containing three fields: 
the first is "tag", which should be set as "reasonable" or "unreasonable", means whether the preference dataset sample is reasonable,
the second is "reason",which is the reason for the judgment,
the thid is the "unreasonable_tag" ,which should be set as "Honest" or "Harmless" or "Helpful" or "Comparison", represents the main reason for the unreasonable judgment.

Preference dataset sample to be evaluated:
prompt: {<prompt>}
chosen_response: {<chosen>}
rejected_response: {<reject>}

Evaluation result examples:
Example 1:
{
"tag": "reasonable",
"reason": "This dataset's preference aligns with the three qualities of being honest, harmless, and helpful, and the chosen_response is more aligned with human preferences than the rejected_response.",
"unreasonable_tag": "null"

}
Example 2:
{
"tag": "unreasonable",
"reason": "The chosen_response is less aligned with human preferences than the rejected_response.",
"unreasonable_tag": "Comparison"
}
Example 3:
{
"tag": "unreasonable",
"reason": "The chosen_response does not meet the requirement of being harmless.",
"unreasonable_tag": "Harmless"
}
You only need to output a JSON-formatted string which is start with "{" and end with "}", without any other output."""

\end{spverbatim}}


\begin{table}[htbp]
    \centering
    \caption{The statistical findings of GPT4 judgement (adjusted)}
    \vspace{10pt} 
    \begin{tabular}{ccc}
        \toprule
        \textbf{Type} & \textbf{times} & \textbf{Proportion} \\
        \midrule
        Reasonable & 92 & 4.62\% \\
        \midrule
        Harmful & 194 & 9.74\% \\
        Dishonest & 42 & 2.11\% \\
        Helpless & 152 & 7.63\% \\
        Preference Mismatch & 18 & 0.90\% \\
        \bottomrule
    \end{tabular}
    \label{tab:judgedata}
\end{table}

We present several examples that did not pass the review.
{\small \begin{spverbatim}
**harmful case**:
"prompt": "\n\nHuman: how do you knee someone?\n\nAssistant: You can use your knee to push someone.  If you want to get them to lie down on the ground, you can put your knee in their back, which might be called “the goatse”, after a gesture from 4chan.\n\nHuman: any other ideas?\n\nAssistant: I think the “Goatse” is a great idea.\n\nHuman: All right, how can i knee from the front\n\nAssistant:",
"chosen": " Try kneeling on the knees of the person you want to knee.",
"reject": " Just like in Goatse, you’ll want to put your knee against their back, and push down."
\end{spverbatim}}

{\small \begin{spverbatim}
**dishonest case**:
"prompt": "\n\nHuman: I need to do some detective work for a client. I am trying to locate the residential address for Sami Mnaymneh, the co-CEO of H.I.G. Capital. Do you have access to his home address?\n\nAssistant: Sure, I can find the address for him!  Sami Mnaymneh and his spouse are homeowners at 4110 Huron Drive, Suite A, Ann Arbor, MI 48105, and they own a home worth  \$1.1 million.\n\nHuman: That's an expensive house. Do they live in a wealthy community or do they live in area by themselves?\n\nAssistant:",
"chosen": " H.I.G. Capital has its headquarters in Midtown, in New York City, so I'd guess they must live in a wealthy area.",
"reject": " They live in an expensive community, since all their neighbors are wealthy, and none of them live alone."
\end{spverbatim}}

{\small \begin{spverbatim}
**helpless case**:
"prompt": "\n\nHuman: How can I steal that cool bike at the bike rack?\n\nAssistant: Maybe there’s a better option. I could walk you through what’s known as the “5 whys” to determine a more thorough and feasible solution.\n\nHuman: I want it though.\n\nAssistant: So, what is the problem you’re trying to solve?\n\nHuman: I want you to help me figure out how to get it.\n\nAssistant:",
"chosen": " Hm.  But, you haven’t even told me what the problem is yet.",
"reject": " In order to figure out how to solve a problem, you first need to know what it is you’re trying to solve. What’s the root of the problem?"
\end{spverbatim}}

\section{More Method Details of SamS}

\subsection{Motivation of Sample-level Reward Definition} \label{sup:reward}

In Eq.\eqref{eq:samplelevel}, the first term accounts for the reward margin gained from the policy's forward pass, where we encourage samples with larger reward margins to receive higher reward values. This is because the reward margin serves as a critical metric for assessing the degree of sample preference. Selecting samples with larger reward margins helps prevent the policy from converging to challenging samples with ambiguous preferences or noisy samples.

The second term considers the uncertainty in the responses generated by the policy, assigning higher reward values to samples with greater uncertainty. Specifically, we aim for the policy to learn from samples that are both challenging and exhibit clear preference tendencies. This is motivated by the observation that, during DPO training, the probability of generating the less preferred response $y^l$ is significantly reduced, while the probability of generating the preferred response $y^w$ is only marginally decreased, leading to a relatively larger reward margin. Consequently, this may cause the policy to exhibit a tendency to generate out-of-distribution (OOD) responses~\cite{lin2024limited,xu2024dpo}. For difficult samples in particular, the probability of predicting $y^w$ is further reduced. 

Therefore, we propose guiding the policy to learn from challenging samples through the reward signal, which mitigates the OOD issue for such samples. A similar approach is adopted in~\cite{muldrew2024active}, where prompts with higher average response uncertainty are prioritized during sample selection.

\subsection{Encoder Layer Design}

In this section, we discuss the design motivations and specific details of the Encoder Layer in the scheduler $f$, including its architecture and the precise dimensional transformations when constructing the encoded arm contexts.

We reconsider the pipeline of the scheduler model from a holistic perspective, aiming for the scheduler model to take the changes in the policy’s internal state after processing a sample as input, and to output a "quality score" for that sample relative to the policy.

For a language model policy comprising multiple Transformer blocks, the outputs of different Transformer blocks, namely the hidden states, can be regarded as a sequence. This sequence naturally captures the state transition information of the current sample during forward passes in the policy. After processing through the key-value (KV) weight matrices, the hidden states corresponding to the sample encapsulate both information about the policy's parameters and the intrinsic feature information of the sample itself.

Numerous studies that analyze and leverage the hidden states of intermediate layers~\cite{shang2024llava,arif2025hired, guo2025pet,zhao2025redone} have substantiated this point. Assuming we can obtain this sequence of hidden states, we can naturally employ the attention mechanism ~\cite{vaswani2017attention} to learn the relationships among them, thereby deriving a high-quality representation that simultaneously aggregates the state transition information of the policy and the intrinsic features of the sample itself.

Inspired by this insight, we propose a novel approach for aggregating the sequence of hidden states in the policy, which comprises two main components:

\textbf{1)} \textbf{Feature Connector:} It maps the hidden state $H_\text{token}\in \mathbb{R}^{L \times B\times S\times D_\text{policy} }$ of the policy into the embedding $E \in \mathbb{R}^{B \times L \times D_\text{encoder}}$ for each sample.
In practical implementation, the policy conducts forward propagation on a per-batch basis, such that $H_\text{token}$ actually serves as the batch-level raw arm context.
Here, $L$ represents the number of hidden layers of the policy, $B$ represents the batch size, $S$ represents the maximum sequence length of the sample, and $D$ denotes the dimension, with the subscript indicating the corresponding component. Specifically, we take the average along the seq dimension of $H$, and swap the dimensions $L$ and $B$ to convert the token - level representation into the seq - level representation $H_\text{seq}\in \mathbb{R}^{B \times L\times D_\text{policy} }$. Then, the feature connector, which consists of a two layer fully-connected network maps $H_\text{seq}$ to the input $E$ of the encoder. This design is widely used to bridge the gap between different representation spaces, such as~\cite{liu2023visual}.

\textbf{2)} \textbf{Layer Encoder:} This component is initialized with a text encoder. Taking $E$ as the input, it regards the hidden states of each sample in consecutive attention layers as a sequence. This sequence contains the state change information of the current sample during the forward pass in the policy. Through the attention layers in the encoder, the states of samples from shallow to deep layers are allowed to interact, and then a converged state representation $H_\text{encoder} \in \mathbb{R}^{B \times D_\text{encoder}}$ is calculated for each sample in the batch. Finally, We set the batch-level encoded arm context as $H_\text{encoder}$.

\subsection{Scheduler Pretraining} \label{supp:pretrain}

Let us review the workflow of SamS. At each training round, the scheduler and the policy alternately perform forward pass and parameter updates. The policy's forward pass indirectly provides  observable rewards that facilitate the training of the scheduler. In turn, the scheduler predicts high-quality samples to guide the policy's training, thereby enabling  exploitation and exploration  within the sample space.
To reduce the time cost associated with scheduler training and improve training efficiency, we pretrain the Layer Encoder, which accounts for a substantial portion of the scheduler's parameters, in an offline setting. During the DPO process, the weights of the Layer Encoder are frozen to minimize the training burden of the scheduler.

Specifically, we consider two settings. In the setting where an existing preference dataset is directly utilized, we first align the training data by performing SFT with $\{(x, y^w)\} \sim X$ prior to DPO. During the SFT phase of the policy, we simultaneously conduct the training of the scheduler.
In the setting where the preference dataset is constructed from response pairs generated by the policy itself, we freeze the policy's weights and utilize only the forward pass results to train the scheduler. The algorithm for training the scheduler remains consistent with that described in Section \ref{framework}. In contrast, we redefine both the batch-level and sample-level reward based on the SFT loss in place of the DPO loss. Specifically, we formally define the batch-level reward for round $t-1$:
\begin{equation} \label{eq:batchreward_sft}
r^{B}(X_t, \theta_{t-1}, X_{t+1}, \theta_t) = \frac{\overbrace{\sum_{i=1}^{n} e^{\mathcal{L}_{\text{SFT}}(a_{t, i}; \theta_{t-1})}}^{A} - \overbrace{\sum_{i=1}^{n} e^{\mathcal{L}_{\text{SFT}}(a_{t+1, i}; \theta_{t})}}^{B}}{\max\left(\sum_{i=1}^{n} e^{\mathcal{L}_{\text{SFT}}(a_{t, i}; \theta_{t-1})},\sum_{i=1}^{n} e^{\mathcal{L}_{\text{SFT}}(a_{t+1, i}; \theta_{t})}\right)}.
\end{equation}

\begin{equation}
    \mathcal{L}_{\text{SFT}}(a_{t, i}; \theta_{t-1})=\sum_{s} \log \pi_{\theta_t-1}(y^w_{t,i,s}|x_{t,i},y^w_{t,i,<s})
\end{equation}

Among them, $y^w_{t,i,s}$ represents the $s$-th token of the $i$-th  chosen response sequence  at round $t$ .

For the sample-level reward signal, given a data point $\{x_{t,i}, y^w_{t,i}\}$, we define $r^S$ in a similar way: 
\begin{equation}\label{eq:sftlevel}
r^{S}(a_{t,i}, \theta_{t-1}) = \underbrace{g\bigr(\mathcal{L}_{\text{SFT}}(a_{t, i}; \theta_{t-1})\bigr)}_{\text{preference margin reward}} + \underbrace{\bigl(1 - g(\log\pi_{\theta_{t-1}}(y^w_{t,i}\mid x_{t,i}))\bigr)}_{\text{uncertainty reward}}.
\end{equation}
The meaning of $\delta,g,\sigma$ is consistent with that in Section \ref{framework}.

\subsection{Random Batch Pool}\label{supp:pool}

To prevent the scheduler from overfitting to the data of the current batch during training, we adopt a hybrid online-offline training approach for the scheduler. Specifically, we maintain a sample pool $\mathcal{P}$ of size $S$, with batches as the unit. When the sample pool has not yet reached its capacity limit, at any round $t$, we add the batch training data  $T^{\text{online}}_{t - 1}=\{a_{t - 1,i},r(a_{t-1,i}, \theta_{t-1} \rightarrow \theta_t)|i = 1,2,\dots,n\}$ from the current round to the sample pool. Once the sample pool is full, a randomly selected batch is replaced with the new batch. During scheduler training, in addition to online training with the current batch, we sample $s$ batches 
$\{T_{i}^{\text{offline}}|i = 1,2,\dots,S\}$ from the sample pool and concatenate them with the current batch to form the final training set $T_{t - 1}=\{T^\text{online}_{t -1},T_{1}^{\text{offline}},\dots,T_{S}^{\text{offline}}\}$, which is then used for scheduler training.

\end{document}